\documentclass[11pt]{article}
\usepackage[utf8]{inputenc} 
\usepackage[T1]{fontenc}    
\usepackage{lmodern}
\usepackage{url}            
\usepackage{booktabs}       
\usepackage{amsfonts}       
\usepackage{nicefrac}       
\usepackage{microtype}      
\usepackage{float}
\usepackage{enumerate}

\usepackage{mathrsfs}
\usepackage{amsmath}
\usepackage{amssymb}
\usepackage{natbib}
\usepackage{graphicx}
\usepackage{url}
\PassOptionsToPackage{x11names}{xcolor}
\usepackage{xcolor}
\PassOptionsToPackage{algo2e,ruled}{algorithm2e}
\usepackage{algorithm2e}
\usepackage{jmlrutils}
\usepackage{hyperref}
\usepackage{nameref}
\newcommand{\vect}{\mathrm{vec}}

\newcommand{\sign}{\text{sign}}
\newtheorem{fact}{Fact}
\newtheorem{observation}{Observation}
\usepackage{authblk}
\DeclareMathOperator*{\argmin}{arg\,min}
\DeclareMathOperator{\R}{\mathbb{R}}

\usepackage{pdfpages} 

\begin{document}
	
	\title{Theory III: Dynamics and Generalization in Deep
		Networks\footnote{This replaces previous versions of Theory III,
			that appeared on the CBMM site and (much more sparsely) on
			arXiv. }}
	


	\author{\large Andrzej Banburski 
	}
	\author {Qianli Liao}
	\author {Brando Miranda}
	\author{ Tomaso Poggio}
	\author{\\ \small   Lorenzo Rosasco}
	\author{Fernanda De La Torre} 
	\normalsize
	
	\author{\small Jack Hidary}
	\affil{Alphabet (Google) X}
	
	%
	
	\date{}

	\maketitle
	
	\begin{abstract}
		
          {The key to generalization is controlling the complexity of
            the network. However, there is no obvious control of
            complexity -- such as an explicit regularization term --
            in the training of deep networks for classification. We
            will show that a classical form of norm control -- but
            kind of hidden -- is present in deep networks trained with
            gradient descent techniques on exponential-type losses. In
            particular, gradient descent induces a dynamics of the
            normalized weights which converge for $t \to \infty$ to an
            equilibrium which corresponds to a minimum norm (or
            maximum margin) solution. For sufficiently large but
            finite $\rho$ -- and thus finite $t$ -- the dynamics
            converges to one of several margin maximizers, with the
            margin monotonically increasing towards a limit stationary
            point of the flow. In the usual case of stochastic
            gradient descent, most of the stationary points are likely
            to be convex minima corresponding to a 
            constrained minimizer -- the network with normalized
            weights-- which corresponds to vanishing
            regularization. The solution has zero
            generalization gap, for fixed architecture, asymptotically for $N \to \infty$,
            where $N$ is the number of training examples. Our approach
            extends some of the original results of
            Srebro from linear networks to deep networks and provides
            a new perspective on the implicit bias of gradient
            descent. We believe that the elusive complexity control we
            describe is responsible for the puzzling empirical finding
            of good predictive performance by deep networks, despite
            overparameterization.  }
		
	\end{abstract}
	
	\section{Introduction}

	In the last few years, deep learning has been tremendously
	successful in many important applications of machine
	learning. However, our theoretical understanding of deep
	learning, and thus the ability of developing principled
	improvements, has lagged behind.  A satisfactory theoretical
	characterization of deep learning is emerging. It covers the
	following questions that are natural in machine learning
	techniques based on empirical risk minimization (see for
	instance \cite{NiyGir96},  \cite{PogSma2003}: 1) {\it representation power} of deep
	networks 2) {\it optimization} of the empirical risk 3) {\it
		generalization properties} of gradient descent techniques
	--- why the expected error does not suffer, despite the
	absence of explicit regularization, when the networks are
	overparametrized? We refer to the latter as the
	non-overfitting puzzle, around which several recent papers
	revolve (see among others \cite{Hardt2016,
		NeyshaburSrebro2017, Sapiro2017, 2017arXiv170608498B,
		Musings2017}).  This paper addresses the third question. 
	
	We start with recent observations on the dynamical
	systems induced by gradient descent methods used for training
	deep networks and summarize properties of the solutions they
	converge to.  Remarkable results by \cite{2017arXiv171010345S}
	illuminate the apparent absence of ''overfitting” in the
	case of linear networks trained on the exponential loss for binary classification.
	They prove that minimization of loss functions such as the
	logistic, the cross-entropy and the exponential loss yields
	asymptotic convergence to the maximum margin solution for
	linearly separable datasets, independently of the initial
	conditions and without explicit regularization.  In this paper, we
	discuss the case of nonlinear multilayer DNNs in the setting
	of separable data, under exponential-type losses and square
	loss, for several variations of the basic gradient descent
	algorithm. Because of homogeneity of the RELU, deep networks
	can be represented as $f(W;x)=\rho f(V;x)$ where $\rho$ is
	the product of the norms of the weight matrix $W$ at each layer
	and $f(V;x)$ is the network with normalized weights $V_k$
	at layer $k$.
	
	Our {\it main result} is that {\it unconstrained gradient
		descent} over an exponential-type loss -- this is the usual
	training procedure for deep networks -- {\it converges to
		solutions $V_k$ that for long but finite time generalize}
	because they are equivalent to constrained minimization or
	equivalently to regularization schemes (which are stable and
	thus generalize). At the limit, they converge to a minimum
	norm solution which is not stable, does not generalize but may
	perform well (in analogy with pseudoinverse). Other results
	are:
	
	\begin{itemize}
		\item Consider gradient descent algorithms -- such as Lagrange
		multipliers methods -- that minimize the exponential loss while
		enforcing a unit $L_p$ norm constraint on the normalized
		weights. The assumption that separability is reached during gradient
		descent implies convergence of the dynamics of the
                $V_k$ to a limit stationary point of the flow for  any finite, fixed $\rho$.
              \item For $\rho \to \infty$ the stationary points
                coincide with the stationary points of the full
                dynamical system obtained from the Lagrange multiplier
                formulation by minimizing also on $V_k$ {\it and}
                $\rho$ .
		\item These limit stationary points are minimum norm --
		and maximum margin -- solutions.
		\item Standard gradient descent used in training deep networks in the
		unnormalized weights followed by $L_2$ normalization (performed
		after stopping gradient descent) has the same qualitative dynamics
		as the Lagrange method and has the  same  stationary points.
		\item Weight normalization and batch normalization have a similar
		qualitative dynamics and converge to the same stationary points.
	\end{itemize}
	\noindent In the perspective of these theoretical results, we discuss
	experimental evidence around the apparent absence of ``overfitting'',
	that is the observation that the expected classification error does
	not get worse when increasing the number of parameters.

	\section{Deep networks: definitions and properties}
	
	{\it Definitions} We define a deep network with $K$ layers with the
	usual coordinate-wise scalar activation functions
	$\sigma(z):\quad \mathbf{R} \to \mathbf{R}$ as the set of functions
	$f(W;x) = W^K \sigma (W^{K-1} \cdots \sigma (W^1 x))$, where
	the input is $x \in \mathbf{R}^d$, the weights are given by the
	matrices $W^k$, one per layer, with matching dimensions. We use the
	symbol $W$ as a shorthand for the set of $W^k$ matrices
	$k=1,\cdots,K$. For simplicity we consider here the case of binary
	classification in which $f$ takes scalar values, implying that the
	last layer matrix $W^K$ is $W^K \in \mathbf{R}^{1,K_l}$. The labels
	are $y_n\in\{-1,1\}$. The weights of hidden layer $l$ are collected in
	a matrix of size $h_l\times h_{l-1}$.  There are no biases apart form
	the input layer where the bias is instantiated by one of the input
	dimensions being a constant. The activation function in this paper is
	the ReLU activation. The norm we use is the $L_2$ unless we say
	otherwise.
	
	{\it Network homogeneity} For ReLU activations the following
	positive one-homogeneity property holds
	$\sigma(z)=\frac{\partial \sigma(z)}{\partial z} z$.  For the network
	this implies $f(W;x)=\prod_{k=1}^K \rho_k f(V_1,\cdots,V_K; x_n)$,
	where $W_k=\rho_k V_k$ with the matrix norm $||V_k||_p=1$. This
	implies the following property of ReLU networks w.r.t. their
	Rademacher complexity:
	\begin{equation}
	\mathbb{R}_N(\mathbb{F}) = \rho
	\mathbb{R}_N(\tilde{\mathbb{F}}), 
	\label{RadaRELU}
	\end{equation}
	\noindent where $\rho=\rho_1 \dotsm \rho_K$, $\mathbb{F}$ is
        the class of neural networks described above and accordingly
        $\tilde{\mathbb{F}}$ is the corresponding class of normalized
        neural networks $f(V;x)$\footnote{This invariance property of the
          function $f$ under transformations of $W_k$ that leaves the
          product norm the same is typical of ReLU (and linear)
          networks.}.  In the paper we will refer to the product
        $\rho=\prod_{k=1}^K \rho_k$ of the norms of the $K$ weight
        matrices of $f$. Note that
	
	\begin{equation}
	\frac{\partial f(W;x)}{\partial \rho_k} = \frac{\rho}{\rho_k}f(V;x)
	\label{rho}
	\end{equation}
	
	\noindent and that the definitions of $\rho_k$, $V_k$ and $f(V;x)$
	all depend on the choice of the norm used in normalization.
	
	{\it Structural property} The following structural property of the gradient of deep
	ReLU networks is useful (Lemma 2.1 of
	\cite{DBLP:journals/corr/abs-1711-01530}):
	\begin{equation}
	\sum_{i,j} \frac{\partial f(W;x)}{\partial W_k^{i,j}}
	W_k^{i,j}=  f(W;x);
	\label{Lemma2.1}
	\end{equation}
	\noindent for $k=1,\cdot,K$.  Equation \ref{Lemma2.1} can be rewritten
	as an inner product between $W_k$ {\it as vectors !}:
	\begin{equation}
	(W_k, \frac{\partial f(W;x)}{\partial
		W_k}) =   f(W;x)
	\label{StructProp}
	\end{equation}
	\noindent where $W_k$ is here the vectorized representation of the
	weight matrices $W_k$ for each of the different layers\footnote{By taking derivatives of both sides of Equation
		\ref{StructProp}, we obtain  to the following property

		\begin{equation}
		(W_k, \frac{\partial^2  f(W;x)}{\partial
			{W_k}^2}) =   0.
		\label{StructProp2}
		\end{equation}

		From Equation \ref{Lemma2.1}, it follows that the condition
		$\left(\frac{\partial f(W;x)}{\partial W_k}\right)=0$ implies
		$f(x)=0$. In the case of square loss, this condition restricts the
		non-fitting stationary points of the gradient to be either linear combinations
		$\sum_{n=1}^N (f(W;x_n)-y_n) \frac{\partial f(W;x)}{\partial W_k}=0$, with
		$\frac{\partial f(W;x)}{\partial W_k} \neq 0, \forall k$ or $f(W;x)=0$. A
		similar restriction also holds for the exponential loss (see Equation
		\ref{standardynamics}).}. The same property holds for $V_k$
	
	\begin{equation}
	(V_k, \frac{\partial f(V;x)}{\partial
		V_k}) =   f(V;x)
	\label{StructPropV}
	\end{equation}

	{\it Gradient flow} The
	gradient flow of the empirical risk $L$  is often written as 
	
	\begin{equation}\dot W\equiv\frac{dW}{dt}=-\gamma(t) \nabla_W(L(f)),
	\end{equation}

	\noindent where $\gamma(t)$ is the learning rate (in this paper we will 
	neglect it). We are well
	aware that the continuous formulation and the discrete one are not equivalent but we are happy to leave a
	careful analysis -- especially of the discrete case -- to better
	mathematicians.
	
	We conjecture that the hypothesis of smooth activations
	is just a technicality due to the necessary conditions for existence
	and uniqueness of solutions to ODEs. Generalizing to differential
	inclusions and non-smooth dynamical systems should allows for these
	conditions to be satisfied in the Filippov sense
	\cite{arscott1988differential}. The Clarke subdifferential is
	supposed to deal appropriately with functions such as RELU.
	
	{\it Separability} When $y_n f(W;x_n) >0$ $\forall n=1,\cdots, N$ we say
	that the data are separable wrt $f \in \mathbf{F}$, that is they can
	all be correctly classified. We assume in this paper that there exist
	$T_0$ such that for $t>T_0$ gradient descent attains a $f$ that
	separates the data. Notice that this is a strong condition on the data
	if $f$ is linear but it is a weak assumption in the case of
	overparametrized, nonlinear, deep networks. In fact here we 
	assume that the condition of {\it separability is reached during
		gradient descent} by the networks we consider.
	

	\section{Related work}
	
	There are many recent papers studying optimization and generalization
	in deep learning. For optimization we mention work based on the idea
	that noisy gradient descent \cite{DBLP:journals/corr/Jin0NKJ17,
		DBLP:journals/corr/GeHJY15, pmlr-v49-lee16, s.2018when} can find a
	global minimum.  More recently, several authors studied the dynamics
	of gradient descent for deep networks with assumptions about the input
	distribution or on how the labels are generated. They obtain global
	convergence for some shallow neural networks
	\cite{Tian:2017:AFP:3305890.3306033, s8409482,
		Li:2017:CAT:3294771.3294828, DBLP:conf/icml/BrutzkusG17,
		pmlr-v80-du18b, DBLP:journals/corr/abs-1811-03804}. Some local
	convergence results have also been proved
	\cite{Zhong:2017:RGO:3305890.3306109,
		DBLP:journals/corr/abs-1711-03440, 2018arXiv180607808Z}. The most
	interesting such approach is \cite{DBLP:journals/corr/abs-1811-03804},
	which focuses on minimizing the training loss and proving that
	randomly initialized gradient descent can achieve zero training loss
	(see also \cite{NIPS2018_8038, du2018gradient,
		DBLP:journals/corr/abs-1811-08888}). In summary, there is by now an
	extensive literature on optimization that formalizes and refines to
	different special cases and to the discrete domain our results of
	Theory II and IIb (see section \ref{BezoutBoltzman}).
	
	For generalization, existing work demonstrate that gradient descent
	works under the same situations as kernel methods and random feature
	methods \cite{NIPS2017_6836, DBLP:journals/corr/abs-1811-04918,
		Arora2019FineGrainedAO}. Closest to our approach -- which is focused
	on the role of batch and weight normalization -- is the paper
	\cite{Wei2018OnTM}. Its authors study generalization assuming a
	regularizer because they are -- like us -- interested in normalized
	margin. Unlike their assumption of an explicit regularization, we show
	here that commonly used techniques, such as batch normalization as
	well as weight normalization, maximize margin while
	controlling the complexity of the classifier without the need to add a
	regularizer or to use weight decay. In fact, we will show that even
	standard gradient descent on the weights  controls the
	complexity of the normalized weights.
	
	Very recently, well after previous versions of this work, two papers
	(\cite{2019arXiv190507325S} and
	\cite{DBLP:journals/corr/abs-1906-05890}) appeared. They develop an
	elegant but complicated margin maximization based approach, describing
	the relations between the {\it margin, the constrained and the
		optimization paths} deriving some of the same results of this
	section (and more). Our approach does not need the notion of maximum margin but
	our theorem \ref{margin-maxTheorem} establishes a connection with it
	and thus with the results of \cite{2019arXiv190507325S} and
	\cite{DBLP:journals/corr/abs-1906-05890}. Our main focus here (and in
	\cite{theory_III}) is on the puzzle of how complexity of deep nets is
	controlled during training despite overparameterization and despite the
	absence of regularization. Our main original contribution is a study
	of the {\it gradient flow of the normalized weights} to characterize
	the implicit {\it control responsible for generalization} in deep networks
	trained under the exponential loss, which describe as implicit $L_2$
	normalization by gradient descent. In the process of doing this, we
	analyze the dynamics of the flow of {\it the direction of the weights}
	induced by gradient descent on the unnormalized weights.

	\section{Main results}
	
	The standard approach to training deep networks is to use stochastic
	gradient descent to find the weights $W_k$ that minimize the empirical
	exponential loss $L= \sum_n e^{-y_nf(W;x_n)}$ by computing
	
	\begin{equation}
	\dot{W_k}  = -\frac{\partial L}{\partial W_k}=
	\sum_{n=1}^N y_n \frac{\partial{f(W; x_n)}} {\partial W_k}  e^{- y_n
		f(W; x_n)} 
	\label{standardynamics}
	\end{equation}
	
	\noindent on a given dataset $\{x_i, y_i\} \quad \forall i=1,\dots,N$.
	
	In this section we study three related
	versions of this problem:
	\begin{enumerate}
		\item the minimization of $L= \sum_n e^{-\rho
			y_nf(V;x_n)}$ under the constraint $||V_k||=1$ wrt $V_k$ for fixed $\rho$;
		\item the minimization of $L= \sum_n e^{-\rho
			y_nf(V;x_n)}$ under the constraint $||V_k||=1$ wrt $V_k, \rho$;
		\item the minimization of $L= \sum_n e^{-\rho
			y_nf(V;x_n)}= \sum_n
		e^{-y_nf(W;x_n)}$ wrt $V_k, \rho$, which is the standard situation for deep nets.
	\end{enumerate}
	
	The section is organized as follows. We will show that problem 1) above
	converges to a stationary point for any finite $\rho$ and then that
	for $\rho \to \infty$ the stationary points of the flow of 1) are the same as the minima of
	2) for $t \to \infty$. Asymptotically these stationary points  are margin
        maximizers and
	correspond (see Appendix \ref{Lorenzo} ) to minimum norm solutions. We
	will then prove that the gradient descent system associated with 3)
	has the same stationary equilibria as problem 2). Finally, we observe
	that for any finite $t$ the solution is regularized, as in early
	stopping for linear networks regression \cite{rosasco2015learning} (for
	$(\lambda(\rho(t)) N)^{-\frac{1}{2}} ~ \epsilon$).  ; for
	$t \to \infty$ the solutions are minimum norm minimizers, a situation
	similar to the case of the pseudoiverse for the linear regression case
	\cite{rosasco2015learning}.

	\subsection{Constrained minimization of the exponential loss}
	
	Generalization bounds suggest constrained optimization of the
	exponential loss that is to minimize $L= \sum_n e^{-\rho
		y_nf(V;x_n)}$ under the constraint $||V_k||=1$ which leads to minimize
	
	\begin{equation}
\mathcal{L}= \sum_n e^{-\rho y_nf(V;x_n)} + \sum_k \lambda_k ||V_k||^2
	\label{ConstrainedOptimization}
	\end{equation}
	
	\noindent  with $\lambda_k$ such that  the constraint $||V_k||=1$ is satisfied.
	
	\subsection{Fixed $\rho$: stationary points of the gradient flow}
	
	Gradient
	descent on $\mathcal{L}$ for fixed $\rho$ wrt $V_k$ yields the dynamical system 
	
	\begin{equation}
	\dot{V_k}= \rho \sum_n e^{-\rho y_nf(V;x_n)}y_n({\frac {\partial f (V;x_n)}
		{\partial V_k}} - V_k f(V;x_n))
	\label{ConstrainedDyn}
	\end{equation}
	\noindent because $\lambda_k= \frac{1}{2} \rho \sum_n e^{-\rho
		f_V(x_n)} f_V(x_n)$,  since
              $\sum_{i,j}(V_k)^{i,j}(\dot{V_k})_{i,j}=0$ because
              $||V_k||^2=1$.

To shorten the expressions that will appear multiple times, let us define
$$
S_k = I-V_k V_k^T,
$$
so that the dynamical system becomes
$$
\dot{V_k}= \rho \sum_n e^{-\rho y_nf(V;x_n)}y_n S_k {\frac {\partial f (V;x_n)}
	{\partial V_k}}.
$$
Let us consider $L(V_k)=\sum_n e^{-\rho y_nf(V;x_n)}$ assuming that
(after convergence of GD) $||V_k||=1$.	
	Since for fixed $\rho$ the domain is compact, minima and maxima $\frac{\partial L}{\partial V_K}=0$ of the
	constrained optimization problem must exist for each fixed $\rho$. Assuming
	data separation is achieved, they satisfy
	
	\begin{equation}
	\sum_n e^{-\rho y_n f(V;x_n)} \frac {\partial f(V;x_n)}
	{\partial V_k}=\sum_n e^{-\rho y_nf(V;x_n)}  V_k f(V;x_n)
	\label{stationarypoints}
	\end{equation}
	
Of course the minimum of the exponential loss $L$ is only zero for the
limit $\rho=\infty$; for any finite $\rho$ the minimum of $L$ is at
the boundary of the compact domain. For any finite, sufficiently large
$\rho$ the minimum is  critical point of the gradient with a positive  semidefinite
Hessian  (see Appendix \ref{CriticalPoints}). In general it is not unique; it is unique  in the case of  one hidden layer linear
networks.
	
\subsubsection{Detailed Analysis}
For $\rho$ sufficiently large, choose the two $x_n$ for which $f(V;x_n)$ is
smallest. Let us call them $x_1$ and $x_2$ with $f(x_1) <f(x_2)$. The
assumption of letting $\rho$ increase to infinity is in practice
equivalent to set $\rho$ to a value
such that $e^{-\rho y_n f(V;x_2)}$ corresponds to zero at machine precision
(whatever it is). For this value of $\rho$ then the  stationary point
satisfies
\begin{equation}
\frac {\partial f (V;x_1)}
	{\partial V_k}=V_k f(V;x_1).
\end{equation}

What if the dynamical system is run with a smaller, fixed $\rho$ than
the ``infinity'' $\rho$ assumed above? In this case, there may be or
not 
	
	\subsection{$\rho \to \infty$ has same stationary points as the full
		dynamical system}
	
	Consider the limit of $\rho \to \infty$ in Equation
	\ref{stationarypoints} .  The asymptotic stationary points of the flow of $V_k$
	then satisfy
	
	\begin{equation}
	\sum_n e^{-\rho y_nf(V;x_n)} y_n S_k\frac {\partial f_V (x_n)}
	{\partial V_k}=0
	\label{AsymptStationaryPoints}
	\end{equation}
	
	\noindent also in the limit $\lim_{\rho \to \infty} $, that is for any
	large $\rho$. So the stationary $V_k$ points  for any large $\rho=R$
	satisfy
	
	\begin{equation}
	\sum_n e^{-R y_nf(V;x_n)} y_n(\frac {\partial f (V;x_n)}
	{\partial V_k} - V_k f(V;x_n))=0.
	\label{AsymptCond}
	\end{equation}

	Consider now gradient descent for the full system obtained
	with Lagrange multipliers, that is, on $\mathcal{L}= \sum_n e^{-\rho y_nf(V;x_n)} +
	\sum_k \lambda_k ||V_k||^2$ wrt $V_k$ {\it and} $\rho_k$, with
	$\lambda_k$ chosen (as before) to implement the unit norm
	constraint. The full gradient
	dynamical system is 
	
	\begin{equation}
	\begin{split}
	\dot{\rho_k}&= \frac{\rho}{\rho_k} \sum_n e^{-\rho y_nf(V;x_n)} y_n
	f(V;x_n) \\
	\dot{V_k}&= \rho \sum_n e^{-\rho  y_nf(V;x_n)}y_n S_k{\frac {\partial f(V;x_n)}
		{\partial V_k}}
	\label{FullDynamics}
	\end{split}
	\end{equation}
	
	Observe (see Appendix) that after onset of separability
	$\dot{\rho_k}>0$ with $\lim_{t \to \infty} \dot{\rho_k}=0$,
	$\lim_{ \to \infty} \rho(t)=\infty$ (for one layer
	$\rho \propto \log t$; for more layer it is faster, see Appendix
	\ref{RHOdynamics}). Appendix \ref{RHOdynamics} shows that $\rho(t)$ is
	a monotonically increasing function from $t=0$ to $t=\infty$ and that
	$\rho_k$ grow at the same rate, independently of the layer $k$. Thus
	for any large $R$ in Equation \ref{AsymptCond}, there exist $T$ such
	that $\rho(T)=R$. At time $T$ then, the condition for the stationary
	point of the $V_k$ in Equation \ref{FullDynamics} is
	\begin{equation}
	\dot{V_k}= R \sum_n e^{-R y_nf(V;x_n)}y_n S_k{\frac {\partial f(V;x_n)}
		{\partial V_k}}=0
	\end{equation}
	\noindent which coincides exactly
	with Equation \ref{AsymptStationaryPoints}.
	
	Thus the full dynamical system \ref{FullDynamics} in the limit of $t \to \infty$
	converges to the same limit -- if it exists -- as does the dynamical
	system Equation \ref{ConstrainedDyn} for $\rho \to \infty$ (in the sense that they have the same stationary points).

	\subsection{Asymptotic stationary points coincide with  minimal norm
		(maximum margin) minimizers}

Here we show that the limit for the two systems exists, is not trivial
and corresponds to maximum margin/minimum norm solutions.  We start
from Equation \ref{AsymptStationaryPoints}.  Without loss of
generality let us assume that the training data $f(V;x_n)$ are ranked
at $t=T_0$ according to increasing normalized margin, that is
$f(V;x_1) \leq f(V;x_2) \leq \cdots \leq f(V;x_N)$. Let us call
$B_k(V;x_n)=y_n (\frac {\partial f(V;x_n)} {\partial V_k}- V_k
f(V;x_n))$. Then the equilibrium condition of Equation
\ref{AsymptStationaryPoints} becomes
\begin{equation}
\begin{split}
&\sum^N_n e^{-\rho(T_0) y_nf(V;x_n)}B_k(V;x_n) =\\ &e^{-R y_1f(V;x_1)} (B_1 +B_2  e^{-R \Delta_2}+\cdots+B_N  e^{-R \Delta_N})=0
\end{split}
\end{equation}
where $\Delta_1=y_1f(V;x_1)-y_1f(V;x_1) =0$,
$\Delta_2=y_2f(V;x_n)-y_1f(V;x_1) \geq 0$,
$\Delta_3=y_3f(V;x_n)-y_1f(V;x_1) \geq 0$ and all $\Delta_n \geq 0$
increase with $n$.  The equation can be rewritten then as
\begin{equation}
e^{-\rho(T_0) y_1f(V;x_1)} (\alpha_1 B_1 + \alpha_2 B_2  +\cdots+\alpha_N  B_N)=0
\label{StationaryCon2}
\end{equation}
\noindent where the $\alpha_n$ are all positive ($\alpha_1=1$) and decreasing with
$n$.

The left hand side of the stationary point equation has the form
$\epsilon (B_1 + \epsilon' B)=0$, with $B=\sum_{n=2}^{N} \alpha_n B_n$,
with both $\epsilon$ and $\epsilon'$ going to zero for $T_0 \to \infty$.
Thus $\lim_{T_0 \to \infty} \epsilon (B_1 + \epsilon' B) = \lim_{T_0 \to
  \infty} \epsilon B_1=0$ can be satisfied for sufficiently large $T_0$
by $B_1=0$, that is by 
$ \frac {\partial f(V;x_1)} {\partial V_k} - V_k
f(V;x_1) = 0 $. This is the condition in which the stationary
point corresponding to $x_1$ provides the maximum margin. Before that limit is reached,
the solution $V_k$ changes with increasing $\rho(t)$.  Thus the
asymptotic stationary points coincide with maximum margin. The
following lemma \cite{theory_III} shows that the margin is
increasing with $t$, after separability is reached and after a single
support vector dominates:
\begin{lemma}
	There exists a $\rho(T_*)$ such that for $\rho>\rho(T_*)$ the sum $\sum_{n=1}^N e^{-
		\rho y_nf(V;x_n)} \propto e^{-
		\rho y_1f(V;x_1)}$. For $\rho>\rho(T_*)$ then the  margin increases
	$y_1\frac{\partial f(V;x_1)} {\partial t}\ge 0$ (even if $\rho$ is
	kept fixed).
	\label{monotonic positive}
\end{lemma}

	{\it Proof} $y_1f(V;x_1)$ increases monotonically or is
	constant because
	\begin{equation}
	\begin{split}
	&y_*\frac{\partial f(V;x_1)}{\partial t}=\sum_k (\frac{\partial y_1f(V;x_1)}{\partial
		V_k})^T \dot{V_k}=\\ &\sum_k \frac{\rho}{\rho_k^2}  e^{- \rho y_1f(V;x_1)} 
	\left(\left\lVert\frac{\partial{f(V;x_1)}} {\partial V_k}\right\rVert_F^2- {f(V;x_1)} ^2\right).
	\label{monotonic}
	\end{split}
	\end{equation}
	
	\noindent Equation \ref{monotonic} implies
	$\frac{\partial y_1 f(V,x_1)} {\partial t}\ge 0$ because
	$||f(V;x_1)||_F = ||V_k^T
	\frac{\partial{f(V;x_1)}} {\partial V_k}||_F \le
	||\frac{\partial{f(V;x_1)}} {\partial V_k}||_F$,
	since the Frobenius norm is sub-multiplicative and
	$V_k$ has unit norm.

\begin{figure}[ht]
	\vskip 0.2in
	\begin{center}
		\centerline{\includegraphics[width=\textwidth]{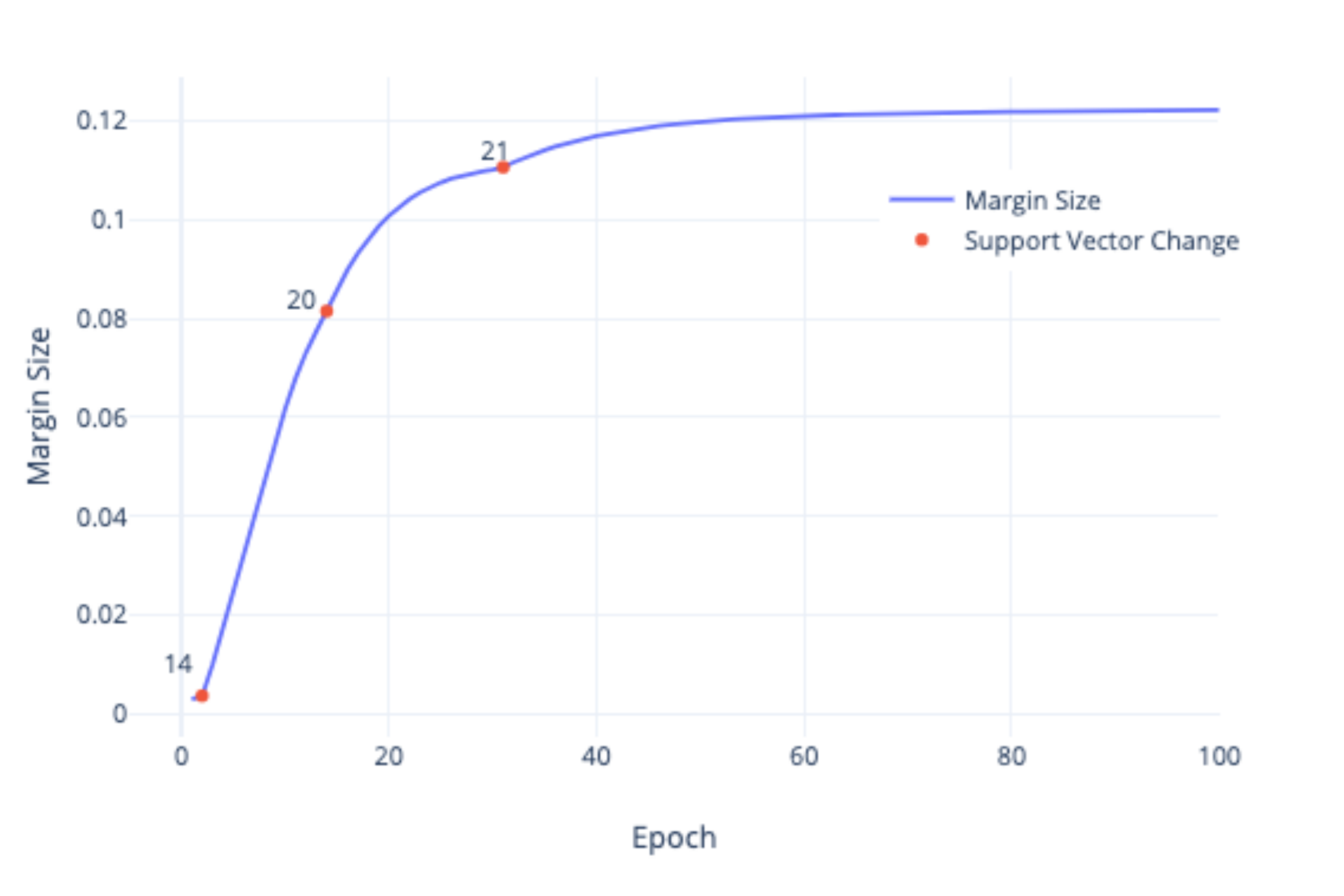}}
		\caption{{\bf Monotonic increase of the margin} The growth of the margin $\min_n y_n f(V;x_n)$ in the binary case. The network converges to 100\% training accuracy in epoch 3 and changes the support vectors three times (red dots -- the numbers above them correspond to the index of the datapoint that becomes the support vector) before stabilizing. As predicted by theory, the margin is non-decreasing. As the support vectors change, notice that the rate of margin growth accelerates.}
		\label{icml-historical}
	\end{center}
	\vskip -0.2in
\end{figure}

We finally note that the maximum margin solution in terms of
$f(V;x)$ and $V_k$ is equivalent to a minimum norm solution in
terms of $W_k$ under the condition of the margin being at least
$1$. This is stated in the following lemma (see Appendix):
\begin{lemma}
	
	The maximum margin problem
	\begin{equation}\label{maxmarg}
	\max_{W_K, \cdots, W_1}\min_{n} y_n f(W; x_n), 
	\quad \text{subj. to}\quad \|W_k\|= 1, \quad
	\forall k.
	\end{equation}
	is equivalent to
		\begin{equation}\label{minnorm}
	\min_{W_k} \frac 1 2 \|W_k\|^2,  \ \text{subj. to}\  y_n f(W; x_n)\ge 1, \quad \forall k, \ n=1, \dots, N.            
	\end{equation}
	
\end{lemma}

	\subsection{Unconstrained gradient descent}
	\label{nounitnorm}
	
	Empirically it appears that GD and SGD converge to solutions that can
	generalize even without any explicit capacity control such as a
	regularization term or a constraint on the norm of the weights. How is
	this possible?  The answer is provided by the fact -- trivial or
	surprising -- that the unit vector $\frac{w(T)}{||w(T)||_2}$ computed
	from the solution $w(T)$ of gradient descent $\dot{w}=-\nabla_w L$ at
	time $T$ is the same, irrespectively of whether the constraint
	$||v||_2=1$ is enforced during gradient descent. This confirms Srebro
	results for linear networks, extending some of them to the deep
	network case. It also throws some light on the nature of the implicit
	bias or hidden complexity control. We show this result next.
	
	
	%
	%

	\subsubsection{Reparameterization of standard gradient descent}
	\label{reparam}
	
	We study the new dynamical system induced by  the dynamical system in
	$\dot{W_k}^{i,j}$ under the reparameterization  $W^{i,j}_k = \rho_k
	V^{i,j}_k$ with $||V_k||_2=1$.
	This is equivalent to changing coordinates from $W_k$ to
	$V_k$ and $\rho_k = ||W_k||_2$. For simplicity of notation we
	consider here for each weight matrix $V_k$ the corresponding
	``vectorized'' representation in terms of vectors $W_k^{i,j} = W_k$.

	We  use the following definitions and properties (for a vector $w$):
	
	\begin{itemize}
		\item The norm $||\cdot||$ is assumed in this section to be the $L_2$ norm.
		\item Define $\frac{w}{\rho}=v$; thus $w=\rho v$ with
		$||v||_2=1$ and $\rho=||w||_2$.
		\item The following relations are easy to check:
		\begin{enumerate}
			\item $\frac{\partial ||w||_2}{\partial w}=v$
			\item Define $S={I-vv^T}=I- \frac {w w^T}{||w||_2^2}$. $S$ has at most
			one zero eigenvalue since $vv^T$ is rank $1$ with a single
			eigenvalue $\lambda=1$.  This means also $S \ge 0$, as can be seen
			directly.
			
			\item $\frac{\partial v}{\partial
				w}=\frac{S}{\rho }$. 
			\item $Sw=S v=0$
			\item $S^2=S$
			\item In the multilayer case, $\frac{\partial{f(x_n; W)}} {\partial
				W_k} = \frac{\rho}{\rho_k}\frac{\partial{\tilde{f}(V; x_n)}} {\partial V_k}$
			\label{Relations}
		\end{enumerate}
	\end{itemize}
	
	The unconstrained gradient descent dynamic system used in training deep networks for the
	exponential loss is given in Equation \ref{standardynamics}, that is 
	
	\begin{equation}
	\dot{W_k}  = -\frac{\partial L}{\partial Wk}=
	\sum_{n=1}^N y_n \frac{\partial{f(W; x_n)}} {\partial W_k}  e^{- y_n
		f(W; x_n)} .
	\end{equation}
	
	Following the chain rule {\it for the time derivatives}, the dynamics  for $W_k$ induces the following dynamics for
	$||W_k||=\rho_k$ and $V_k$:
	\begin{equation}
	\dot{\rho_k}= \frac{\partial ||W_k||}{\partial W_k} \frac{\partial
		W_k}{\partial t}= V_k^T \dot{W_k}
	\end{equation}
	\noindent and
	\begin{equation}
	\dot{V_k}= \frac{\partial V_k}{\partial W_k} \frac{\partial
		W_k}{\partial t}= \frac {S_k}{\rho_k} \dot{W_k}
	\label{vdot}
	\end{equation}
	\noindent where $S_k= I- V_kV_k^T$.
	We now obtain the time derivatives of $V_k$ and $\rho_k$ from the time
	derivative of $W_k$; the latter is computed from the 
	gradients of $L$ with respect to $W_k$ that is from the gradient
	dynamics of $W_k$.  Thus {\it unconstrained gradient descent}
	coincides with the following dynamical system
	\begin{equation}
	\dot{\rho_k}= V_k^T \dot{W_k}=\sum_{n=1}^N V_k^T y_n \frac{\partial f(W;x_n)} {\partial W_k}  e^{- y_n
		f(W;x_n)}= \frac{\rho}{\rho_k} \sum_{n=1}^N  y_n  e^{-
		\rho y_nf(V;x_n)} y_n f(V;x_n)
	\label{wdot1}
	\end{equation}
	
	\noindent and 
	
	\begin{equation}
	\dot{V_k}=\frac{\rho}{\rho_k^2}\sum_{n=1}^N  e^{- \rho y_n  f(V;x_n)} y_n
	(\frac{\partial{f(V;x_n)}} {\partial V_k}-V_k f(V;x_n)),
	\label{wdot2}
	\end{equation}
	
	\noindent where we used the structural lemma to set $V_k  V_k^T \frac{\partial{\tilde{f}(x_n)}} {\partial
		V_k} = V_k \tilde{f}(x_n)$.
	
	
	Clearly the dynamics of {\it unconstrained gradient
		descent} and the dynamics of {\it constrained
		gradient descent} are very similar
	since they differ by a $\rho^2$ factor in the
	$\dot{v}$ equations. The conditions
	for the stationary points of the gradient for the $v$
	vectors -- that is the values for which $\dot{v}=0$ --
	are {\it the same in both cases} since for any $t>0$
	$\rho(t)>0$.
	
	\subsubsection{Constrained optimization and weight normalization}
	
	We recall that {\it constrained gradient descent} using Lagrange
	multipliers yields the dynamical system
	
	\begin{equation}
	\begin{split}
          \dot{\rho_k}&= \frac{\rho}{\rho_k} \sum_n e^{-\rho y_n f(V;x_n)}
          y_n f(V;x_n) \\
          \dot{V_k}&= \rho \sum_n y_n e^{-\rho y_n f(V;x_n)}({\frac {\partial f (V;x_n)} {\partial V_k}} - V_k f(V;x_n)).
          \end{split}
	\end{equation}
	
	Constrained normalization  by tangent gradient gives the same
	dynamical system, as expected. The Appendix shows that it coincides with the
	so-called weight normalization algorithm. 
	
\subsection{Linear networks}

In the case of linear networks under the exponential loss, we use
stochastic gradient descent to find the weights $w$ that minimize
the empirical exponential loss $L= \sum_n e^{-y_nw^T x_n}$ by computing
	
	\begin{equation}
	\dot{w}  = -\frac{\partial L}{\partial w}=
	\sum_{n=1}^N y_n x_n e^{- y_n
		w^T x_n} 
	\label{standardynamicsLinear}
	\end{equation}

We assume linear separability, that is $w^T x_n y_n >0$. The dynamical
system 	\ref{standardynamicsLinear} diverges with $||w|| \to
\infty$. We consider instead the gradient dynamics of the system with
 Lagrange multipliers and $L= \sum_n e^{-\rho y_n v^T x_n}+\lambda ( ||v||^2-1)$

	\begin{equation}
	\dot{\rho}=  \ \sum_{n=1}^N  e^{-
		\rho y_nv^Tx_n} y_nv^Tx_n
	\end{equation}
	
	\noindent and 
	
	\begin{equation}
	\dot{v}=- \frac{\partial L}{\partial v} = \rho \sum_{n=1}^N  e^{- \rho y_n v^T x_n} 
	y_nx_n- 2 \lambda v  .
	\end{equation}

The value of $\lambda$ that ensures  $||v||=1$ is
$\lambda=\frac{\rho}{2}\sum_n e^{- \rho y_nv^T x_n} y_nv^T x_n$. The
critical points correspond to $\dot{v}=0$ that is to $\rho \to \infty$
or if a set of  support vectors with value $f(w;x^*)$ is reached at a
finite time to 
\begin{equation}
		x^*- vv^T x^*=0
	\end{equation}
\noindent which gives $v=x^*$ for the value of the weights at the
critical point which has to be a minimum of the loss and a maximum of
the margin (see earlier).

The Hessian of $L$ (and Jacobian of $L$)  is given by

$$H^{ij} = \sum_n e^{-\rho y_nv^T x_n} \rho^2 x_n^i x_n^j + 2\lambda \delta^{ij}.$$
If we now plug in the value of $\lambda$, we get

$$H^{ij} = \sum_n e^{-\rho y_nv^T x_n} (\rho^2 x_n^i x_n^j + \rho y_nv^T x_n \delta^{ij})$$

Since $x_n x_n^T$ is positive semi-definite, and we have separability,
the Hessian of the loss is  positive definite and, in particular, it
is positive definite at the critical point $v=x^*$ .

%
%
%

	\subsection{Main result}
        Actual convergence for the constrained case at a finite time
        happens in the case of a set of support vectors with the same
        margin 	$\sum^N_n e^{-R y_nf(V;x_n)} y_n(\frac {\partial f(V;x_n)}
	{\partial V_k} - V_k f(V;x_n))= (\frac {\partial f (V;x_*)} {\partial V_k} - V_k f(V;x_*)) =
        0 $ is valid, which corresponds to a large but finite $\rho$
        and a small but non-zero $\lambda$. The unconstrained case has
        the same solutions.  There are obvious limitations to this
        asymptotic statement. In fact, it turns out that the dynamics
        we described converges for a certain $N$ to a minimum norm,
        maximum margin solution similar to a pseudoinverse which does
        not technically generalize (expected error equal to empirical
        error) for that value of $N$ but can perform well in terms of
        expected error. In addition, while an appropriately normalized
        network $f(V;x)$ can have a small generalization gap at
        some finite $N$ under the exponential loss, bounds on the
        classification error for the same finite $N$ remain an open
        problem. More importantly, the case $\sum^N_n e^{-R y_nf(V;x_n)}=
        e^{-\rho y_*f(V;x_*)}$ may never happens during the finite times
        of a gradient descent run.

	\begin{figure*}[t!]\centering
		\includegraphics[trim = 0 90 0 0, width=1.0\textwidth, clip]{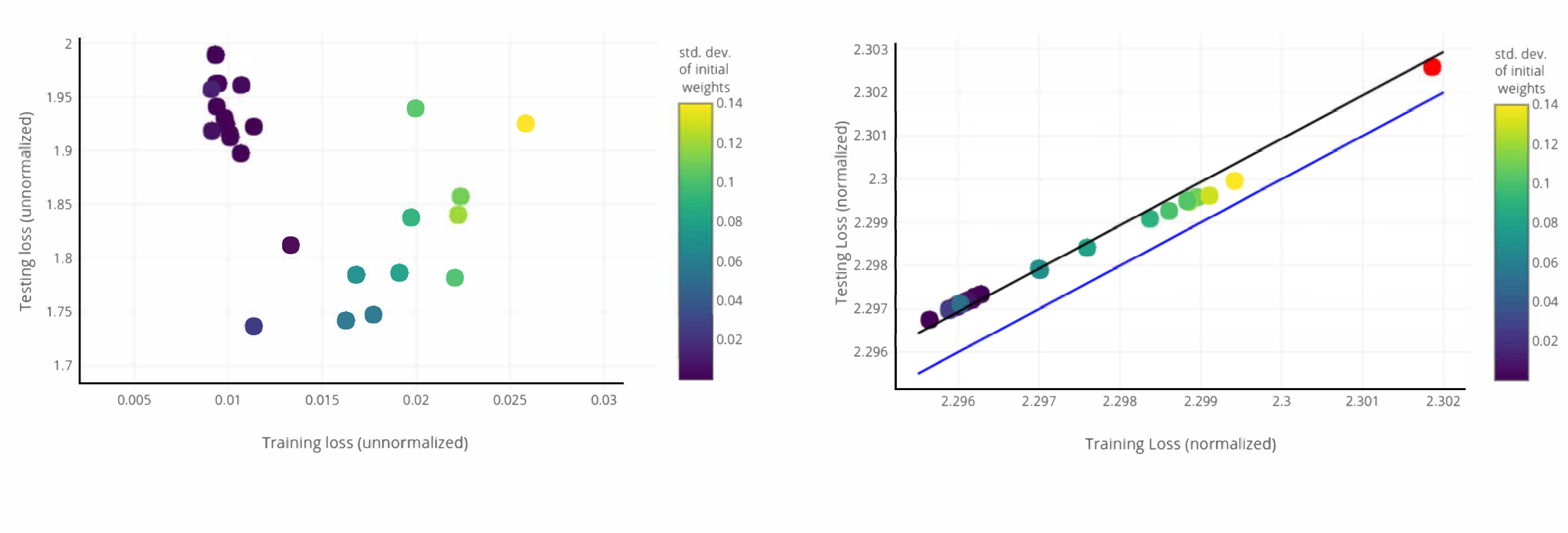} 
		\caption{\it Empirical evidence of generalization by
			normalized networks with respect to the cross
			entropy loss. The left graph shows testing vs
			training cross-entropy loss for networks each
			trained on the same data sets (CIFAR10) but with 
			different initializations, yielding zero
			classification error on training set but different
			testing errors. The right graph shows the same
			data, that is testing vs training loss for the same
			networks, now normalized by dividing each weight by
			the Frobenius norm of its layer. Notice that all
			points have zero classification error at
			training. The red point on the top right refers to a
			network trained on the same CIFAR10 data set but
			with randomized labels. It shows zero classification
			error at training and test error at chance
			level. The top line is a square-loss regression of
			slope $1$ with positive intercept. The bottom line
			is the diagonal at which training and test loss are
			equal.  The networks are 3-layer convolutional
			networks. The left can be considered as a
			visualization of generalization bounds when the
			Rademacher complexity is not controlled. The right
			hand side is a visualization of the same relation
			for normalized networks that is
			$L(f) \leq \hat{L}(f) +
			c_1\mathbb{R}_N(\mathbb{\tilde{F}}) + c_2 \sqrt{
				\ln(\frac{1}{\delta})/2N}$. Under our
			conditions for $N$ and for the architecture of the
			network the terms
			$c_1\mathbb{R}_N(\mathbb{\tilde{F}}) + c_2 \sqrt{
				\ln(\frac{1}{\delta})/2N}$ represent a small
			offset. 
		}
		\label{main}
	\end{figure*}

	\subsection{Summary}
	
	The following theorem summarizes our main results

\begin{theorem}
	\label{Theorem23}
	Assume that separability is reached at time $T_0$ during gradient
	descent on  the exponential loss, that is $y_n f(W;x_n) >0, \quad \forall n$. Then
	unconstrained gradient descent converges in terms of the
	normalized weights to a solution that is under complexity
	control for any finite time. In
	addition the following  properties hold:
	\begin{enumerate}
		\itemsep-0.2em 
		\item Consider the dynamics (A) resulting from using Lagrange
		multipliers on the constrained optimization problem: ``minimize
		$L= \sum_n e^{-\rho y_nf(V;x_n)}$ under the constraint
		$||V_k||=1$ wrt $V_k$''. The dynamics converges for any fixed $\rho$
		to stationary points of the $V_k$ flow that are 
                minima with  a positive semi-definite Hessian.
		\item Consider the dynamics (B) resulting from using Lagrange
		multipliers on the constrained optimization problem: ``minimize
		$L= \sum_n e^{-\rho y_nf(V;x_n)}$ under the constraint
		$||V_k||=1$ wrt $V_k$ and $\rho_k$''. 
		The  stationary
		points of $V_k$ in (B) in the limit of  $t \to \infty$  coincide with the limit 
		$\rho \to \infty$ in the dynamics (A) and they are maxima of the margin.
		\item The unconstrained gradient descent dynamics converges to the same
		stationary points of the flow of $V_k$ as (A) and (B).
		\item Weight normalization\cite{SalDied16} corresponds to dynamics (B).
		\item For each layer $\frac{\partial
			\rho_k^2} {\partial t}$ is the same irrespectively of $k$.
		\item In the 1-layer network case
		$\rho  \approx \log t$ asymptotically. For deeper
		networks,  the product of weights at each layer diverges faster than
		logarithmically, but each individual layer diverges slower than in the
		1-layer case
		\item Gradient flow converges for infinite time to
                  directions given by a  set of support vectors with the
		same   value $y_1f(V;x_1)$ that satisfy the set of $K$ coupled vector
		equations $ (\frac {\partial f (V;x_1)} {\partial V_k} = V_k
		f(V;x_1))$.

	\end{enumerate}
\end{theorem}

	In the Appendix \ref{LinearNets} we describe analysis of convergence
	for linear networks.

	\section{Discussion}
	
	Our main results are {\it for classification} in the setting of {\it
		separable data} for  {\it  continous 
		gradient flow} under {\it the exponential loss}.
	
	The key assumption is that at some point during gradient descent
	separability is reached. The assumption is reasonable because of
	overparameterization and empirically satisfied for many data
	sets. However, an analysis of the dynamical system before separability
	is lacking so far and should be the goal of further work.  Our main
	result is that there is an {\it implicit $L_2$ norm constraint on the
		$V_k$ dynamics in standard gradient descent for deep networks with
		RELUs.} Therefore standard gradient descent on the weights, provides
	a solution $f(V;x)$ that generalizes without the need of additional
	explicit regularization or explicit norm constraints.
	
	To reach that conclusion we establish several properties of the
	dynamical system that are interesting in their own. As a starting
	point, the Appendix proves, under the assumption of separability, that
	for each layer $\frac{\partial \rho_k^2} {\partial t}$ grows
	monotonically with $t$, at the same rate irrespectively of $k$. In the
	1-layer network case $\rho \approx \log t$ asymptotically. For deeper
	networks, the product of weights at each layer diverges faster than
	logarithmically, but each individual layer diverges slower than in the
	1-layer case.  Then, the dynamics of the normalized weights in {\it
		the standard unconstrained gradient descent} on the exponential loss
	has {\it the same asymptotic stationary points as gradient descent on
		the regularized loss (with vanishing $\lambda$)}. Furthermore, the
	equilibrium point of the flow of $V_k$ for $\rho \to \infty$
	corresponds to a minimum norm, maximum margin solution of the constrained
	optimization problem.
	
	Notice that gradient descent methods minimizing an empirical loss with
	a $L_p$ norm constraint implement a classical recipe for good
	predictivity. They converge to stationary points of the gradient of
	$V_k$ which attain zero loss -- assuming separability occurs during
	gradient descent -- for $\rho \to \infty$. The dynamics of $W_k$ for
	separable data attain the global zero minimum of the loss for
	$||W_k|| \to \infty$ but diverges in $W$. The dynamics of the
	direction of the weights $V_k$ in a multilayer network under
        SGD is likely to have several convex
	minima for any finite $\rho$ which is sufficiently large.
	
	Examples of techniques commonly used to train over-parametrized,
	multilayer, RELU, deep networks are {\it weight normalization} and
	{\it batch normalization} enforcing an explicit unit constraint in the
	$L_2$ norm of $V_k$. For linear networks (and for one support vector)
	the convergence rate of standard GD ($\frac{1}{\log t}$) is slower
	than the convergence rate ($\frac{1}{t^{\log (\sqrt(t)}}$) of weight
	normalization.

	The fact that the solution corresponds to a maximum margin (or minimum
	norm) solution  may explain the following puzzling
	behavior (see  Figures in the Supplementary Material, in which batch
	normalization was used): the test classification error does not get
	worse when the number of parameters increases well beyond the number
	of training data. This may be  because the dynamical system is trying to
	maximize the {\it margin} under unit norm of $f(V;x)$.

	The gradient flows corresponding to normalization with
	different $L_p$ norms are expected to be different and to converge to
	different solutions, as in the classical case of support vector
	machines with $L_2$ vs $L_1$ regularizers.  It is useful to emphasize
	that despite the similarities between some of the methods enforcing
	unit constraints in the 2-norm, they usually correspond to different
	dynamical flows but with the same qualitative dynamics and the same
	stationary solutions. In particular, batch normalization and weight
	normalization do not have the same dynamics; in turn they are slightly different
	from standard gradient descent. Furthermore, our analysis has been
	restricted to the continuous case; the discrete case may yield greater
	differences. An additional remark  is that {\it weight
		normalization and batch normalization are enforcing explicitly a
		constraint (on the dynamics of $V_k$) which is natural for obtaining
		generalization}, and is deeper than reducing covariate shifts (the
	properties described in \cite{2018arXiv180511604S} are fully
	consistent with our characterization in terms of a norm
	constraint). Notice that as in the
	case of the vanishing regularizer assumed in the original theorem of
	\cite{DBLP:conf/nips/RossetZH03}, the {\it Lagrange multipliers
		$\lambda_k$ should go to zero fast enough for $\lambda_k \rho_k$ to
		go to zero while $\rho_k \to \infty$.}

The basic complexity control mechanism we uncovered--
regularization -- explains the asymptotic generalization behavior (for $N \to
\infty$) of deep networks but does not explain the more common regime
of $N<<D$ in which 
overparametrized deep networks fit the training data and
perform well on out-of-sample points.  It is therefore useful to recall that the classical
analysis of Empirical Risk Minimization (ERM) algorithms studies their
asymptotic behavior for the number of data $N$ going to infinity. In
this limiting regime, $N>D$ where $D$ is the fixed number of weights;
consistency (informally the expected error of the empirical minimizer
converges to the best in the class) and generalization (the empirical
error of the minimizer converges to the expected error of the
minimizer) are equivalent. The capacity control described in this note
implies that {\it there is asymptotic generalization and consistency in
deep networks} for $n \to \infty$ and fixed architecture.

The non-asymptotic behavior in the overparametrized regime is similar
to the regression case of linear kernel methods
\cite{2018arXiv180800387L,2018arXiv181211167R,2019arXiv190805355M,Misha2019}). This
phenomenon suggests that under certain conditions, the
pseudoinverse\footnote{It is not well known, but easy to verify using
  the Mathlab function ``cond'', that the condition number associated
  with a random data matrix is usually worse for $N=D$, better for
  $N>D$ and even better for $N<<D$. The proof is probably well-know to
  the experts since we found versions of it on the margins of a few
  linear algebra books. The implication is that data errors are
  amplified most when $N=D$ in the estimation of new data; in an
  equicalent way the ratio of out-of-sample prediction and
  in-sample-prediction is largest for $N=D$, better for $N>D$ and best
  for $N<<D$.} may perform well in terms of expected error while the
generalization gap (difference between expected and empirical loss) is
large. We note that is not surprising that complexity control is a
prerequisite for good performance even in an overparametrized regime
in which the classical analysis via generalization does not apply. The
pseudoinverse solution is unique and continuous, satisfying the key
conditions -- including stability with respect to noise in the data --
of a well-posed problem. We also remark that the analysis of the
dynamics of deep networks described in this paper, once adapted to the
square loss, suggests that the weights $W_k$ of each layer converge to
local minimum norm minimizers, because of the iterative regularization
properties of gradient descent (in analogy with the fully linear case
\cite{rosasco2015learning}).

%

	In addition, commonly used
	weight decay with appropriate parameters can induce generalization
	since it is equivalent to regularization. Furthermore, typical
	implementations of data augmentation may also effectively avoid
	overparameterization: at each iteration of SGD only ``new'' data are
	used and depending on the number of iterations it is quite possible
	that the size of the training data exceeds the number of
	parameters. Within this online framework, one expects convergence to
	the minimum of the expected risk (see Supplementary Material section
	on Data Augmentation) without the need to invoke generalization
	bounds.
	
	There are of course several open problems. It seems that under certain
	conditions neural networks can be described in terms of a ``neural tangent
	kernel''  in a linear way (wrt weights) -- as hinted in
	several recent papers (see for instance \cite{2018arXiv180406561M} and
	\cite{2019arXiv190202880N}). This regime, corresponding to large norm
	initializations, is also characterized by low accuracy in terms of the
	expected error. Can we explain this behavior in terms of our
	formalization of the dynamics? A more general question is of key
	interest also for applications: can we characterize the conditions
	that ensure convergence to the ``best'' of the maximum margin
	solutions?  Less important but still interesting is the question of
	why batch normalization is empirically better than weight
	normalization? This requires some explanation because both enforce
	implicitly or explicitly a unit constraint in the $L_2$ norm.
	
	\small
	\subsubsection*{Acknowledgments}
	We thank Yuan Yao, Misha Belkin, Jason Lee and especially Sasha
	Rakhlin for illuminating discussions. Part of the funding is from
	Center for Brains, Minds and Machines (CBMM), funded by NSF STC award
	CCF-1231216, and part by C-BRIC, one of six centers in JUMP, a
	Semiconductor Research Corporation (SRC) program sponsored by DARPA.
	
	\normalsize
	\small
	\newpage
	\bibliographystyle{unsrt}
	\bibliography{Boolean}
	\normalsize
	\newpage
	\begin{center}
		{\bf \large Appendix}
	\end{center}

	\section{The optimization landscape of (unnormalized) Deep RELU Networks under
		exponential-type loss}
	\label{BezoutBoltzman}
	
	The {\it first part} of the argument of this section relies on the
	simple observation  that RELU
	networks, under the hypothesis of an exponential-type loss function,
	do not have zeros of the gradient (wrt the $W_k$) that separate the data. In fact, under the
	hypothesis of an exponential-type loss, separable data and homogeneity
	of the network -- such as kernel machines and deep RELU networks --
	the only stationary points of the gradient that separate the data
	are for $\rho=\infty$.
	
	Notice that minima arbitrarily close to zero loss exist for any
	finite, large $\rho$. For $\rho \to \infty$, the Hessian becomes
	arbitrarily close to zero, with all eigenvalues being close to zero. On
	the other hand, any point of the loss at a finite $\rho$ has a
	Hessian wrt $W_k$ which is not identically zero: for instance in the linear case the Hessian is proportional
	to $\sum_n^N x_n x^T_n$.  
	
	Consider now that the local minima which are not global minima must
	missclassify. How degenerate are they?  In the case of a linear
	network in the exponential loss case, assume there is a finite $w$ for
	which the gradient is zero in some of its components. One question is
	whether this is similar to the regularization case or not, that is
	whether {\it misclassification regularizes}.

	Let us look at a linear example:
	
	\begin{equation}
	\dot{w} = F(w) = - \nabla_{w} L(w)= \sum_{n=1}^n y_nx_n^T e^{- y_nx_n^T w} 
	\end{equation}
	
	\noindent in which we assume that there is one classification
	`        error (say for $n=1$), meaning that the term $e^{- y_1x_1^T w}$
	grows exponentially with $w$. Let us also assume that gradient
	descent converges to $w^*$. This implies that
	$\sum_{n=2}^n y_nx_n^T e^{- y_nx_n^T w^*}= -y_1x_1^T e^{- y_1x_1^T w^*}$:
	for $w^*$ the gradient is zero and $\dot{w}=0$. Is this a
	convex equilibrium?  Let us look at a very simple $1D$,
	$n=2$ case: 
	
	\begin{equation}
	\dot{w}= - x_1 e^{x_1 w^*}  + x_2 e^{-x_2 w^*}
	\end{equation}
	
	\noindent If $x_2>x_1$ then $\dot{w}=0$ for $ e^{(x_1+x_2)w^*}=\frac{x_2}{x_1}$
	which implies $w^*= \frac{\log(\frac{x_2}{x_1})}{x_1+x_2}$. This is clearly a hyperbolic equilibrium point, since we have
	
	\begin{equation}
	\nabla_w F(w) = -x_1^2 e^{x_1 w^*}  - x_2^2 e^{-x_2 w^*} < 0,
	\end{equation}
	\noindent so the single eigenvalue in this case has no zero real part.
	
	In general, if there are only a small number of classification errors, one
	expects a similar situation for some of the components. {\it
		Differently from the regularization case, misclassification
		errors do not ``regularize'' all components of $w$ but only
		the ones in the span of the misclassified examples}.
	
	The more interesting case is with $D >N$. An example of this
	case is $D=3$ and $N=2$ in the above equation. The Hessian wrt
	$W_k$ at the minimum will be degenerate with at least one zero
	eigenvalue ad one negative eigenvalue. 

	The stationary points of the gradient of $f$ in the nonlinear multilayer
	separable case under exponential loss are given by

	\begin{equation}
	\sum_{n=1}^N y_n \frac{\partial{f(W;x_n)}} {\partial W^{i,j}_k}  e^{- y_n
		f(W;x_n)} = 0.
	\end{equation}
	
	This means that 
	the global zeros of the loss are at infinity, that is for
	$\rho \to \infty$ in the exponential. If other stationary points were to
	exist for a value $W^*$ of the weights, they would be given by
	zero-valued linear combinations with positive coefficients of
	$\frac{\partial f(W;x_n)} {\partial W^{i,j}_k}$. Use of the
	structural Lemma shows that
	$\frac{\partial f(W;x)} {\partial W^{i,j}_k}=0, \forall i,j,k$
	implies $f(W^*;x)=0$. So stationary points of the gradient wrt $W_k$ that are
	data-separating do not exist for any finite $\rho$. The situation is quite
	different if we consider {\it  stationary points wrt $V_k$}.

	Clearly, it would be interesting to characterize better the degeneracy of
	the local minima. For the goals of this section however the fact that
	they cannot be completely degenerate is sufficient. We thus have the
	following rather obvious result:
	
	\begin{theorem}
		Under the exponential loss, the weight  $W_k$ for zero loss at
		infinite $\rho$ are completely
		degenerate, with all eigenvalues of the Hessian being zero. The other
		stationary points of the gradient are less degenerate, with at least
		one  nonzero eigenvalue.
	\end{theorem}

	The {\it second part} of our argument (in \cite{theory_IIb}) is that SGD
	concentrates on the most degenerate minima. The argument is based 
	on the fact that the Boltzman distribution is formally the asymptotic ``solution'' of the
	stochastic differential Langevin equation and also of SGDL, defined as
	SGD with added white noise (see for instance
	\cite{raginskyetal17}. More informally, there is a certain
	similarity between SGD and SGDL suggesting that in practice the
	solution of SGD may be similar to the solution of SGDL. The Boltzman distribution is
	
	\begin{equation}
	p(W^{i,j}_k) = \frac{1}{Z}e^{-\frac{L(f)}{T}},
	\label{Bolzman}
	\end{equation}
	
	\noindent where $Z$ is a normalization constant, $L(f)$ is the loss
	and $T$ reflects the noise power. The equation implies that SGDL
	prefers degenerate minima relative to non-degenerate ones of the same
	depth. In addition, among two minimum basins of equal depth, the one
	with a larger volume is much more likely in high dimensions as shown
	by the simulations in \cite{theory_IIb}. Taken together, these two
	facts suggest that SGD selects degenerate minimizers corresponding to
	larger isotropic flat regions of the loss. Then SDGL shows concentration --
	{\it because of the high dimensionality} -- of its asymptotic
	distribution Equation \ref{Bolzman}.
	
	Together \cite{theory_II} and \cite{theory_IIb} imply the following 
	
	{\bf Conjecture}  {\it For 
		overparametrized deep networks under an exponential-type  loss, SGD selects with high
		probability global minimizers of the empirical loss, which are fully 
		degenerate (for $\rho \to \infty$, in the separable case.}

	\section{Uniform convergence bounds: minimizing a surrogate loss under  norm constraint}
	\label{Early stopping}
	
	Classical {\it generalization bounds for regression}
	\cite{Bousquet2003} suggest that minimizing the empirical loss of a
	loss function such as the cross-entropy
	subject to constrained {\it complexity of the minimizer} is a way to
	to attain  generalization, that is an expected loss close to the
	empirical loss:
	
	\begin{theorem}
		The following generalization bounds 
		apply to $\forall f \in \mathbb{F}$ with probability at least
		$(1-\delta)$:
		\begin{equation}
		L(f) \leq \hat{L}(f) + c_1\mathbb{R}_N(\mathbb{F}) + c_2 \sqrt
		\frac{\ln(\frac{1}{\delta})}{2N}
		\label{bound}
		\end{equation} 
	\end{theorem}
	\vskip0.1in
	\noindent where $L(f) = \mathbf E [\ell(f(x), y)]$ is the expected
	loss, $\hat{L}(f)$ is the empirical loss, $\mathbb{R}_N(\mathbb{F})$
	is the empirical Rademacher average of the class of functions
	$\mathbb{F}$ measuring its complexity; $c_1, c_2$ are constants that
	depend on properties of the Lipschitz constant of the loss function,
	and on the architecture of the network.
	
	Thus minimizing under a constraint on the Rademacher complexity a
	surrogate function such as the cross-entropy (which becomes the
	logistic loss in the binary classification case) will minimize an
	upper bound on the expected classification error because such
	surrogate functions are upper bounds on the $0-1$ function\footnote{
		Furtermore the excess classification risk $R(f)-R^*$, where $R(f)$ is
		the classification error associated with $f$ and $R^*$ is the Bayes
		error \cite{Bartlett03convexity}, can be bounded by a monotonically
		increasing function of appropriate surrogate functions such as the
		exponential and the cross-entropy.}. Calling $\rho \tilde{f}=f$,  using the
	homogeneity of the network, one can use the
	following version of the bound above:
	
	\begin{theorem}
		$\forall f \in \mathbb{F}$ with probability at least
		$(1-\delta)$:
		\begin{equation}
		L(\rho\tilde{f}) \leq \hat{L}(\rho\tilde{f}) + \rho \mathbb{R}_N(\mathbb{\tilde{F}}) + c_2 \sqrt
		\frac{\ln(\frac{1}{\delta})}{2N}
		\end{equation} 
	\end{theorem}
	\vskip0.1in
	\noindent and to use $\rho$ effectively as a parameter.
	%
	%
	%

	In this setup, $\tilde{f}$ is obtained by minimizing the exponential loss for
	$\rho \to \infty$ under a unit norm constraint on the weight matrices
	of $\tilde{f}$:
	\begin{equation}
	\lim_{\rho \to \infty} \arg\min_{||V_k||=1, \  \forall k} L(\rho \tilde{f}) 
	\label{UnitNormMin}
	\end{equation} 
	As it will become clear later, gradient
	descent techniques on the exponential loss automatically increase
	$\rho$ to infinity.

	In the following we explore the implications for deep networks of this classical
	approach to generalization.

	\section{Constrained minimization of the exponential loss
		implies margin maximization }
	\label{Rosset-TP}
	Though not critical for our approach to the question of generalization
	in deep networks it is interesting to observe that constrained
	minimization of the exponential loss implies margin maximization. This
	property relates our approach to the results of several recent papers
	\cite{2017arXiv171010345S,
		2019arXiv190507325S,DBLP:journals/corr/abs-1906-05890}. Notice that
	our theorem \ref{margin-maxTheorem} as in
	\cite{DBLP:conf/nips/RossetZH03} is a {\it sufficient condition for margin
		maximization}. Sufficiency is not true for general loss functions.  In fact \cite{
		2019arXiv190507325S} seems to require additional conditions for
	ensuring that the margin path converges to the optimization path.
	
	To state the margin property more formally, we adapt to our setting a
	different result due to \cite{DBLP:conf/nips/RossetZH03} (they
	consider a vanishing $\lambda$ regularization term whereas we have a
	unit norm constraint). First we recall the definition of the empirical
	loss $L(f)=\sum_{n=1}^N \ell(y_n f(x_n))$ with an exponential loss
	function $\ell(yf)= e^{-yf}$. 
	We define $\eta(f)$
	a the {\it margin} of $f$, that is $\eta(f)=\min_n f(x_n)$.
	
	Then our margin maximization theorem  takes the form
	
	\begin{theorem} 
		Consider the set of $V_k, k=1,\cdots, K$ corresponding to
		
		\begin{equation}
		\min_{{||V_k||}=1} L(f(\rho_k, V_k))
		\label{V(rho)}
		\end{equation} 
		\noindent where the norm $||V_k||$ is a chosen $L_p$ norm and
		$L(f)(\rho_k, V_K) = \sum_ n \ell(y_n \rho f(V; x_n))$ is the
		empirical exponential loss. For each layer $k$ consider a sequence of increasing
		$\rho$. Then the associated sequences of $V_k$ defined by Equation
		\ref{V(rho)}, converges for $\rho_k \to \infty$ to the maximum
		margin of $f$, that is to
		$\max_{||V_k|| \leq 1} \eta(f)$ .
		\label{margin-maxTheorem} 
	\end{theorem}

	This means that for $\rho \to \infty$ the weights $V_k$ that minimize the exponential loss
	under a unit norm constraint converge to the weights $V_k$ that maximize
	the margin of $f(V;x)$. Thus
	\begin{equation}
	\argmin_{{||V_k||=1}, \rho} L(\rho f) (V;x_n) \to \arg\max_{||V_k|| = 1} \min_n y_nf(V;x_n)
	\end{equation}  
	
	{\it Proof} The proof loosely follows
	\cite{DBLP:conf/nips/RossetZH03,2017arXiv171010345S}, see also
	\cite{Wei2018OnTM}. We observe that $\min_{{||V_k||}=1} L(f(\rho))$
	exists because $L(f(\rho))$ is continuous since $f$ is continuous and
	the domain is compact for any finite $\rho_k$.  We carry on the
	argument for a specific $k$ with the understanding that the same steps
	should be done for each $k$. In the following we drop $k$ for
	simplicity of notation. We now assume that $V^*$
	minimize $L(f(\rho))$. We claim that the associated network
	$f^*$ maximizes the margin for $\rho \to \infty$. Arguing by
	contradiction assume that for a given $\rho$ there exist a different
	$f_1$ with a larger margin, that is
	$\eta(f_1) > \eta(f^*)$ . Then we can choose any
	$f_2$ with weights $V_2$ such that $||V_2-V^*|| \leq \delta$
	such that $\eta(f_2) < \eta(f_1) - \epsilon$. Now we
	can choose a $\rho$ large enough so that $\rho f_1$ has a
	smaller loss then $\rho f_2$, implying that $f^*$
	cannot be a convergence point. The last step is based on the fact that
	{\it if $\eta(f_1) > \eta(f_2)$, then
		$L(\rho f_1) < L(\rho f_2)$ for large enough
		$\rho$. }
	
	This follows from the existence of $\rho$ such that
	$L(\rho y_1f_1) \leq N e^{-\rho \eta(f_1)} \leq e^{-\rho\eta(f_2)} \leq L(\rho y_2f_2)$.

	
	Theorem \ref{margin-maxTheorem} can be supplemented  with the following
	lemma, proved in the following Appendix \ref{Lorenzo}:
	
	\begin{lemma}
		Margin maximization of the network with normalized weights is
		equivalent to norm minimization under strict separability ($y_n f(x_n)
		\ge 1$).
		\label{LorenzoLemma}
	\end{lemma}

	\section{Minimal norm and maximum margin}
	\label{Lorenzo}
	We discuss the connection between maximum margin and minimal
        norms problems in binary classification.  To do so, we reprise
        some classic reasonings used in the context of support vector
        machines. The norm assumed in this section is the $L_2$ norm
        (the proof can be extended to other norms).  We assume
        functions with the one-homogeneity property, namely, for all
        $\alpha>0$, 	$$f(\alpha W; x)= \alpha f( W; x)$$. We also
        assume separability, that is $y_n f(x_n) >0, \quad \forall n$.
	
	Given a training set of $N$ data points $(x_i, y_i)_{i=1}^N$,
        where labels are $\pm 1$,  the functional margin is  defined as
	\begin{equation}\label{fmarg}
	\min_{i=1, \dots, N} y_i f(W_K \cdots W_1; x_i).
	\end{equation}

	The maximum (max) margin problem is then 
	\begin{equation}\label{maxmarg}
	\max_{W_K, \cdots, W_1}\min_{i} y_i f(W_K, \cdots, W_1; x_i), 
	\quad \quad \quad \text{subj. to}\quad \|W_k\|= 1, \quad
        \forall k.
	\end{equation}

	The latter constraint is needed to avoid trivial solutions in
        light of the one-homogeneity property.  We next show that

\begin{theorem}
        Problem~\eqref{maxmarg} is equivalent to
	\begin{equation}\label{minnorm}
	\min_{W_k} \frac 1 2 \|W_k\|^2, \quad \quad \forall k \quad \text{subj. to}\quad  y_i f(W_K, \cdots, W_1; x_i)\ge 1, \quad \quad i=1, \dots, N.
	\end{equation}
\end{theorem}

	To see this, we introduce a number of equivalent formulations. 
	First,  notice that functional margin~\eqref{fmarg} can be equivalently written as 
	$$
	\max_{\gamma>0 } \gamma, \quad \quad \quad \text{subj. to}\quad 
	y_i f(W_K, \cdots, W_1; x_i)\ge\gamma, \quad \quad i=1, \dots, N.
	$$
	Then, the max margin problem~\eqref{maxmarg}  can be written as 
	\begin{equation}\label{maxmarg2}
	\max_{W_K, \cdots, W_1, \gamma>0 } \gamma, 
	\quad \quad \quad \text{subj. to}\quad \|W_k\|= 1, \quad
        \forall k \quad y_i f(W_K, \cdots, W_1; x_i)\ge\gamma, \quad \quad i=1, \dots, N.
	\end{equation}
	Next, we can incorporate the norm constraint noting that using one-homogeneity, 
	$$
	y_i f(W_K, \cdots, W_1; x_i)\ge\gamma \Leftrightarrow 
	y_i f(\frac{W_K}{\| W_K \|}, \cdots, \frac{W_1}{\| W_1 \|} ;
        x_i)\ge \gamma \rho \Leftrightarrow 
	y_i f(W_K, \cdots, W_1; x_i) \ge\gamma'
	$$
	\noindent so that Problem~\eqref{maxmarg2} becomes
	\begin{equation}\label{maxmarg3}
	\max_{W_K, \cdots, W_1; \gamma'>0 } \frac{\gamma'}{\rho}, 
	\quad \quad \quad \text{subj. to}\quad \quad y_i f(W_K, \cdots, W_1; x_i)\ge\gamma', \quad \quad i=1, \dots, N.
	\end{equation}
	Finally, using again  one-homogeneity,  without loss of generality, we can set $\gamma'=1$ and obtain the equivalent problem
	\begin{equation}\label{minnorm2}
	\max_{W_K, \cdots, W_1} \frac{1}{\rho}, 
	\quad \quad \quad \text{subj. to}\quad \quad y_i f(W_K, \cdots, W_1; x_i)\ge1, \quad \quad i=1, \dots, N.
	\end{equation}
	The result is then clear noting that 
	$$
	\max_{W} \frac{1}{\rho}\Leftrightarrow 
	\min_{||W_K||, \cdots, ||W_1||} \rho \Leftrightarrow
	\min_{\rho_1,\cdots,\rho_K} \frac{\rho^2}{2}.
	$$

	\section{Gradient techniques for norm control} 
	\label{normcontroltechniques}
	
	There are several ways to implement the minimization in the tangent
	space of $||V||^2=1$.  In fact, a review of gradient-based algorithms
	with unit-norm constraints \cite{845952} lists
	
	\begin{enumerate}
		\item the {\it Lagrange multiplier
			method} 
		\item the {\it coefficient
			normalization method} 
		\item the {\it tangent gradient method} 
		\item the {\it true gradient method} using natural gradient.
	\end{enumerate}
	
	For small values of the step size, the first three techniques are
	equivalent to each other and are also good approximations of the true
	gradient method \cite{845952}. The four techniques are closely related and have the
	same goal: performing gradient descent optimization with a unit norm
	constraint.
	
	{\it Remarks} 
	\begin{itemize}
		\item  Stability issues for numerical implementations are discussed in
		\cite{845952}.
		\item  Interestingly, there is a close relationship between the
		Fisher-Rao norm and the natural gradient
		\cite{DBLP:journals/corr/abs-1711-01530}.  In particular, the natural
		gradient descent is the steepest descent direction induced by the
		Fisher-Rao geometry.
		\item Constraints in optimization such as $||v||_p=1$ imposes a geometric
		structure to the parameter space. If $p=2$ the weight vectors
		satisfying the unit norm constraint form a n-dimensional hypersphere of radius $=1$. If
		$p=\infty$ they form an hypercube. If $p=1$ they form a hyperpolyhedron.
		
	\end{itemize}

	\subsection{Lagrange multiplier method} 
	\label{Lagrange}

	We start with one of the techniques, the Lagrange multiplier method,
	because it enforces the unit constraint in an especially transparent way.  We
	assume separability and incorporate the constraint in the exponential  loss by defining a new
	loss as
	
	\begin{equation}
	L= \sum_{n=1}^N e^{-\rho f(V;x_n) y_n } + \sum_{k=1}^K\lambda_k (||V_k||_p^p-1)
	\label{RegLoss}
	\end{equation} 
	
	\noindent where the Lagrange multipliers $\lambda_k$ are chosen to
	satisfy $||V_k||_p=1$ at convergence or when the algorithm is stopped.

	We perform gradient descent on $L$ with respect to $\rho, V_k$.  We obtain for $k=1,\cdots,K$
	
	\begin{equation}
	\dot{\rho_k} = \sum_n  \frac{\rho}{\rho_k} e^{-\rho(t) f(V; x_n)y_n} y_n f(V;x_n), 
	\label{rhodot}
	\end{equation}
	\noindent and for each layer $k$
	\begin{equation}
	\dot{V_k}  =\rho(t) \sum_n  e^{- \rho(t) y_nf(V;x_n)}y_n \frac {\partial f(V;x_n)}
	{\partial V_k} (t) + \lambda_k(t) p  V_k^{p-1} (t).
	\label{minimumnorm}
	\end{equation}
	
	The sequences $\lambda_k(t)$ must satisfy
	$\lim_{t \to \infty} ||V_k||_p=1 \quad \forall k$.

	{\it Remarks}
	\begin{enumerate}
		
		\item In the case of  $p=2$, with the conditions
		$||V_k||=1$ at each $t$, $\lambda_k(t)$ must satisfy
		\begin{equation}
		||V_k(t) + \rho(t) \sum_n  e^{-\rho (t)y_nf(V;x_n)}y_n \frac {\partial f(V;x_n)} {\partial V_k} - 
		2 \lambda_k(t) V_k(t)||=1.
		\end{equation}
		\noindent  Thus defining $g(t)=\rho(t) \sum_n  e^{- \rho(t) y_nf(V;x_n)} y_n\frac {\partial f(V;x_n)}
		{\partial V_k} (t)$ we obtain 
		\begin{equation}
		||V_k(t) + g(t) +2 \lambda_k(t) V_k(t)||=1,
		\end{equation}
		\noindent that is 
		\begin{equation}
		||\alpha(t) V_k(t) + g(t)||=1,
		\end{equation}
		\noindent with $\alpha(t)= 1+2 \lambda_k(t) $.
		The solution for $\alpha$ is
		\begin{equation}
		\alpha(t)=\sqrt{1-||g(t)||^2+(V_k^T g(t))^2}-V^T(t)
		g(t).
		\label{alpha}
		\end{equation}
		\noindent Thus $\lambda$ goes to zero at infinity because $g(t)$ does
		and $\alpha \to 1$.
		
		\item 
		Since the first term in the right hand side of Equation
		(\ref{minimumnorm}) goes to zero with $t \to \infty$ and the Lagrange multipliers $\lambda_k$ also
		go to zero, the normalized weight vectors converge at infinity to
		$\dot{V}_k=0$. On the other hand, $\rho(t)$ grows to
		infinity. As shown in section \ref{nounitnorm}, the
		norm square $\rho^2_k$ (when $p=2$) of each layer grows at the same rate.
		\item As in the case of the of the vanishing regularizer assumed in
		the original theorem of \cite{DBLP:conf/nips/RossetZH03}, the
		Lagrange multipliers $\lambda_k$ here go to zero.

		\item The Lagrange multiplier approach with $\lambda_k \to 0$
		establishes a connection with  Halpern iterations and minimum norm
		solutions for degenerate minima.
		

	\end{enumerate}

	\subsection{Coefficient  normalization method}
	
	If $u(k)$ is unconstrained the  gradient maximization of $L(u)$ with
	respect to $u$ can be performed using the algorithm
	\begin{equation}
	u(k+1)=u(k) + g(k)
	\end{equation}
	where $g(k)= \mu(k) \nabla_u L$. Such an update, however, does not
	generally guarantee that $||u^T(k+1)||=1$. 
	The coefficient normalization method  employs a two step update
	$\hat{u} (k+1) = u(k) + g(k)$ and $u(k+1)=\frac{\hat{u}(k+1)}{||\hat{u}(k+1)||_p}$.
	
	\subsection{Tangent gradient method}
	
	\begin{theorem} (\cite{845952} ) Let $||u||_p$ denote a vector norm that is
		differentiable with respect to the elements of $u$ and let $g(t)$ be
		any vector function with finite $L_2$ norm.  Then, calling
		$v(t)=\frac{\partial ||u||_p}{\partial u}_{u=u(t)}$, the equation
		\begin{equation}
		\dot{u}=h_g(t)=Sg(t)= (I-\frac{v v^T}{|v||_2^2}) g(t)
		\label{dot_u}
		\end{equation}
		\noindent with $||u(0)|| =1$, describes the flow of a vector $u$ that
		satisfies $||u(t)||_p=1$ for all $t \ge 0$. 
		\label{Theorem1}
	\end{theorem}
	
	In particular, a form for $g$ is  $g(t)= \mu(t) \nabla_u L$, the
	gradient update in a gradient descent algorithm. We call $Sg(t)$ the
	tangent gradient transformation of $g$. For more details see \cite{845952}.
	
	In the case of $p=2$ we replace $v$ in Equation \ref{dot_u} with $u$
	because $v(t)=\frac{\partial ||u||_2}{\partial u}=u$. This gives $S=
	(I-\frac{u u^T}{|u||_2^2})$ and $\dot{u}=Sg(t).$
	
	{\it Remarks}
	
	\begin{itemize}
		\item For $p=2$   $v=\frac{\partial ||u||_p}{\partial u}_u$ is $v=
		\frac{u}{||u||_2}$
		\item For $p=1$,  $\frac{\partial ||u||_1}{\partial
			u_j}=\frac{u_j}{|u_j|}$.
		\item For $p=\infty$, $\frac{\partial ||u||_\infty}{\partial
			u_j}=\sign (u_k) \delta_{k,j}$, if maximum is attained in
		coordinate $k$.
	\end{itemize}

	\section{Standard dynamics, Weight Normalization and Batch Normalization}
	
	We now discuss the relation of some existing techniques for  training deep
	networks with the gradient descent techniques under unit norm
	constraint of the previous section.
	
	\subsection{Standard unconstrained dynamics}
	\label{OurWeightNormalization}
	
	The standard gradient dynamics is given by

	\begin{equation}
	\dot{W}^{i,j}_k  = -\frac{\partial L}{\partial W^{i,j}_k}= \sum_{n=1}^N y_n \frac{\partial{f(W;x_n)}} {\partial W^{i,j}_k}  e^{- y_n
		f(W;x_n)} 
	\end{equation}
	
	\noindent where $W_k$ is the weight matrix of layer $k$.  As
	we observed, this dynamics has global minima at infinity with
	zero loss, if
	the data are separable. The other stationary points have loss
	greater than zero.

	Empirical observations suggest that unconstrained gradient
	dynamics on deep networks converges to solutions that
	generalize, especially when SGD is used instead of GD. In our
	experiments, normalization at each iteration, corresponding to
	the coefficient normalization method, improves generalization
	but not as much as Weight Normalization does.
	
	\subsection{Weight Normalization}
	\label{Weight Normalization}
	
	For each layer (for simplicity of notation and consistency with
	the original weight normalization paper), weight
	normalization \cite{SalDied16} defines $v$ and $g$ in terms of
	$w=g \frac{v}{|v|}$. The dynamics on $g$ and $v$ is induced by the
	gradient dynamics of $w$ as follows (assuming $\dot{w}=-\frac{\partial
		L}{\partial w}$

	\begin{equation}
	\dot{g}=\frac{v^T}{||v||} \dot{w}
	\label{gflow}
	\end{equation}
	
	\noindent and
	
	\begin{equation}
	\dot{v}=\frac{g}{||v||}  S \dot{w}
	\label{vflow1}
	\end{equation}
	
	\noindent with $S=I- \frac{v v^T}{||v||^2}$.
	
	We claim that this is the same dynamics obtained from tangent gradient for $p=2$. In
	fact, compute the flows in $\rho,v$ from $w=\rho v$ as
	
	\begin{equation}
	\dot{\rho}=\frac{\partial w}{\partial \rho} \frac{\partial
		L}{\partial w}= v^T\dot{w}
	\label{rhoflow1}
	\end{equation}
	
	\noindent and

	\begin{equation}
	\dot{v}= S \rho \dot{w}
	\label{v-flow-withunitnorm}
	\end{equation}

\begin{figure*}[h!]\centering
		\includegraphics[width=0.8\textwidth]{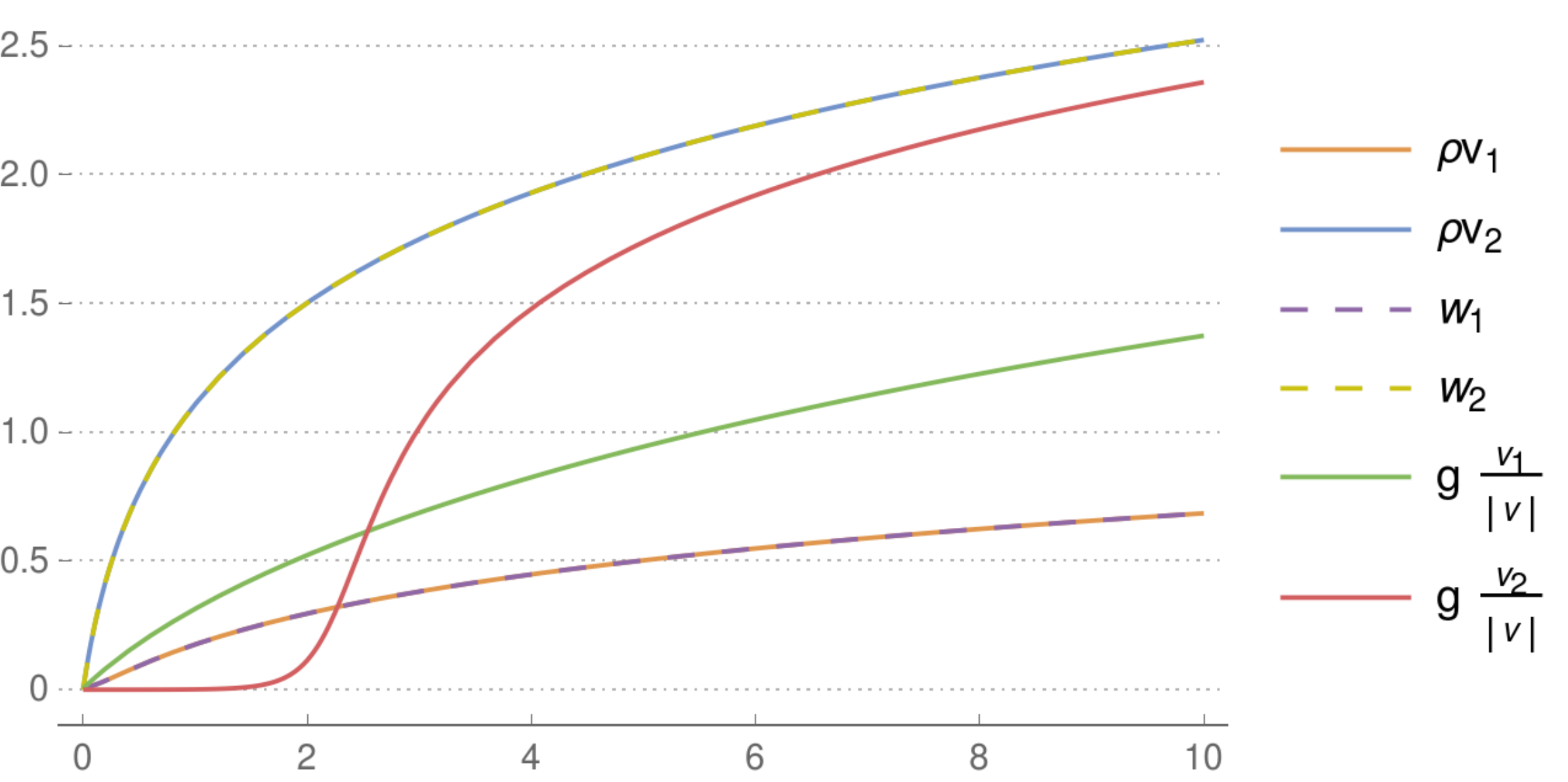}
		\caption{\it Example comparison of dynamics
			$\dot{W_k}=-\frac{\partial L}{\partial W_k}$ (dashed lines)
			and their equivalence to the reparameterization of the
			unconstrained dynamics (orange and blue) as a function of
			time. Standard weight normalization leads to different
			trajectories of gradient descent (in green and red). The
			example is that of a linear network with only two parameters
			and two training examples, using an exponential loss.}
		\label{gradientdynamics}
	\end{figure*}

	Clearly the dynamics of this algorithm is the same as standard weight
	normalization if $||v||_2=1$, because then Equations \ref{gflow} and
	\ref{vflow1} become identical to Equations \ref{rhoflow1} and
	\ref{v-flow-withunitnorm} with $g$ corresponding to $\rho$. We now
	observe, multiplying Equation \ref{vflow1} by $v^T$, that $v^T \dot{v}=0$
	because $v^T S=0$, implying that $||v||^2$ is constant in
	time. Thus if $||v||=1$ at initialization, it
	will not change (at least in the noiseless case). Thus {\it the dynamics of Equations \ref{gflow} and
		\ref{vflow1} is the same dynamics as Equations \ref{rhoflow1} and
		\ref{v-flow-withunitnorm}}.  It is also easy to see that the
	dynamics above is not equivalent to the standard dynamics on the $w$
	(see also Figure \ref{gradientdynamics}).
	
	\subsection{Batch Normalization}
	\label{BatchNormalization}
	
	Batch normalization \cite{ioffe2015batch} for unit $i$ in the network normalizes the
	input vector of activities to unit $i$ -- that is it normalizes $X^j = \sum_j W^{i,j} x_j$, where
	$x_j$ are the activities of the previous layer. Then it sets the
	activity to be
	
	$$Y^j = \gamma \cdot \hat{X}^j + \beta = \gamma \frac{X^j - \mu_B}{\sqrt{\sigma^2_B + \epsilon}} + \beta,$$
	where $\gamma, \beta$ are learned subsequently in the optimization and
	$$ \mu_B = \frac{1}{N}\sum_{n=1}^N X_n \qquad \sigma_B^2 = \frac{1}{N}\sum_{n=1}^N (X_n - \mu_B)^2.$$
	Note that both $\mu_B$ and $\sigma_B^2$ are vectors, so the division
	by $\sqrt{\sigma^2_B + \epsilon}$ has to be understood as a point-wise
	Hadamard product $\odot (\sigma^2_B + \epsilon)^{-1/2}$.  The gradient
	is taken wrt the new activations defined by the transformation
	above. 
	
	Unlike Weight Normalization, the Batch Normalization equations do not
	include an explicit computation of the partial derivatives of $L$ with
	respect to the new variables in terms of the standard gradient
	$\frac{\partial L}{\partial w}$. The reason is that Batch
	Normalization works on an augmented network: a BN module is added to
	the network and partial derivatives of $L$ with respect to the new
	variables are directly computed on its output. Thus the BN algorithm
	uses only the derivative of $L$ wrt the old variables as a function of
	the derivatives of $L$ wrt new variables in order to update the
	parameters below the BN module by applying the chain rule.  Thus we
	have to estimate what BN {\it implies} about the partial derivatives
	of $L$ with the respect to the new variables as a function of the
	standard gradient $\frac{\partial L}{\partial w}$.
	
	To see the nature of the dynamics implied by batch normalization we
	simplify the original Equations (in the Algorithm 1 box in
	\cite{ioffe2015batch}). Neglecting $\mu_B$ and $\beta$ and $\gamma$, we
	consider the core transformation as
	$\hat{X} =\frac{X}{\sigma_B}$ which, assuming fixed inputs,
	becomes $\hat{X} = \frac{X}{|X|}$ which is mathematically
	identical with the transfomation of section \ref{OurWeightNormalization} $v =
	\frac{w}{|w|}$. In a similar way the dynamics of $w=\frac{\partial
		L}{\partial  w}$ induces the following dynamics on $\hat{X}$:
	
	\begin{equation}
	\dot{\hat{X}}= \frac{\partial
		\hat{X}}{\partial X} \dot{X} 
	\end{equation}
	\noindent where $\dot{x} = \nabla_x L$. We consider $X\in \mathbb{R}^{N\times D}$. In the $D=1$ case, we get
	$$\frac{\partial \hat{X}}{\partial X} = (\sigma_B^2 + \epsilon)^{-1/2}\left[ - \frac{1}{N}\hat{X}\hat{X}^T + I\right]. $$ 
	In the general $D$-dimensional vector case, this generalizes to
	$$\frac{\partial \hat{X}}{\partial X} = (\sigma_B^2 + \epsilon)^{-1/2}\left[ - \frac{1}{N}\hat{X}^T\odot \hat{X} + I\right]. $$ 
	
	Notice that $I - \hat{X}\hat{X}^T = S$. Since $x = W x_{input}$ this
	shows that batch normalization is closely related to {\it gradient
		descent algorithms with unit $L_2$ norm constraint of the tangent
		gradient type}.  Because of the simplifications we made, there are
	other differences between BN and weight normalization, some of which
	are described in the remarks below.
	
	{\it Remarks}
	\begin{enumerate}
		
		\item Batch normalization (see Supplementary Material), does not control
		directly the norms of $W_1, W_2, \cdots, W_K$ as WN
		does. Instead  it controls the norms
		\begin{equation}
		||x||, ||\sigma(W_1 x)||, ||\sigma(W_2 \sigma(W_1x))||, \cdots
		\end{equation}
		\noindent In this sense it implements a somewhat weaker version of the
		generalization bound.
		\item In the multilayer case, BN  controls separately the norms $||V_i||$
		of the weights into unit $i$, instead of controlling the overall Frobenius
		norm of the matrix of weights as WN does. Of course control of the
		$||V_i||$ implies control of $||V||$ since $||V||^2=\sum_i ||V_i||^2_i$.
		
	\end{enumerate}

	\subsection{Weight Normalization and Batch Normalization enforce an
		explicit unit 2-norm constraint}
	
	Consider the {\it tangent gradient transformation} to a gradient
	increment $g(t)= \mu(k) \nabla_u L$ defined as $h_g=S g(t)$ with
	$S=I-\frac{u u^T}{||u||_2^2}$.  Theorem
	\ref{Theorem1} says that the dynamical system $\dot{u}=h_g$ with
	$||u(0)||_2 =1$ describes the flow of a vector $u$ that satisfies
	$||u(t)|_2=1$ for all $t \ge 0$. It is obvious then that
	
	\begin{observation} The dynamical system describing weight
		normalization ( Equations \ref{gflow} and \ref{vflow1}) 
		are not changed by the tangent gradient transformation.
		\label{NoChange}
	\end{observation}
	
	The proof follows easily for WN by using the fact that $S^2=S$. The same
	argument can be applied to BN. The property is consistent with the
	statement that they enforce an $L_2$ unit norm constraint.
	
	Thus all these techniques implement margin maximization of $f_V$
	under unit norm constraint of the weight matrices of
	$f_V$. Consider for instance the Lagrange multiplier method. Let
	us assume that starting at some time $t$, $\rho(t)$ is large enough
	that the following asymptotic expansion (as $\rho\rightarrow\infty$)
	is a good approximation:
	$\sum_n e^{- \rho(t) y_nf(V;x_n)} \sim C\max_n e^{- \rho(t)
	y_nf(V;x_n)} $, where $C$ is the multiplicity of the $x_n$ with
	minimal value of the margin of $f(V;x_n)$. The data points with the corresponding
	minimum value of the margin $y_nf(V;x_n)$ are called support
	vectors (the  $x_i,y_i$ s.t  ${\arg \min_ny_nf(V;x_n)}$). They are a subset of cardinality $C$ of the $N$ datapoints,
	all with the same margin $\eta$. In particular, the term
	$g(t)=\rho(t) \sum_n e^{-\rho (t)y_nf(V;x_n)}y_n \frac {\partial
	f(V;x_n)} {\partial V_k}$ becomes
	$g(t)\approx \rho(t) e^{-\rho (t)\eta} \sum^C_i y_i\frac {\partial
		f(V;x_i)} {\partial V_k}$.
	
	As we mentioned, in GD with unit norm constraint there will be
	convergence to $\dot{V}_k=0$ for $t \to \infty$. There may be
	trajectory-dependent, multiple alternative selections of the support
	vectors (SVs) during the course of the iteration while $\rho$ grows:
	each set of SVs may correspond to a max margin, minimum norm solution
	without being the global minimum norm solution. Because of Bezout-type
	arguments \cite{theory_II} {\it we expect multiple maxima}. They
	should generically be degenerate even under the normalization
	constraints -- which enforce each of the $K$ sets of $V_k$ weights to
	be on a unit hypersphere. Importantly, the normalization algorithms
	ensure control of the norm and thus of the generalization bound even
	if they cannot ensure that the algorithm converges to the globally
	best minimum norm solution (this depends on initial conditions for
	instance).

	\section{Dynamics of $\rho$}
	\label{RHOdynamics}
	
	First we show that for each layer $\frac{\partial
		\rho_k^2} {\partial t}$ is the same irrespectively of $k$. Then we
	consider the dynamics of $\rho$. In the 1-layer network case
	the $\rho$ dynamics yields $\rho  \approx \log t$ asymptotically. For deeper
	networks,  we will show
	that the product of weights at each layer diverges faster than
	logarithmically, but each individual layer diverges slower than in the
	1-layer case. 
	%

	\subsection{The rate of growth of $\rho_k$ is the same for all layers}
	A property of the dynamics of $W_k$, shared
	with the dynamics of $V_k$ under explicit unit norm constraint, is suggested by
	recent work \cite{NIPS2018_7321}: {\it the rate of change of the squares of
		the Frobenius norms of the weights of different  layers is the same
		during gradient descent}. This implies that if the weight matrices are
	small at initialization, the gradient flow corresponding to
	gradient descent maintains approximatevely equal Frobenius norms
	across different layers, which is {\it a consequence  of the norm
		constraint}. This property is expected in a minimum norm situation,
	which is itself equivalent to maximum margin under unit norm (see
	Appendix \ref{Lorenzo}). The observation of \cite{NIPS2018_7321} is
	easy to prove in our framework. 
	%
	Consider the gradient descent equations 
	
	\begin{equation}
	\dot{W}^{i,j}_k  = \sum_{n=1}^N y_n [\frac{\partial{f(W;x_n)}} {\partial W^{i,j}_k}]  e^{- y_n
		f(W;x_n)}.
\label{GDdyn}
	\end{equation}
	
We use 
	the relation $\dot{||W_k||}=\frac{\partial ||W_k||}{\partial W_k}
	\frac{\partial W_k}{\partial t}= \frac{W_k}{||W_k||}
        \dot{W_k}$ by multiplying both sides of Equation \ref{GDdyn}
        by $W_k$, summing over $i,j$ and using lemma
        \ref{Lemma2.1}. Thus we obtain the following dynamics on $||W_k||$ :
	
	\begin{equation}
	\dot{||W_k||}  = \frac{1}{||W_k||} \sum_{n=1}^N y_n f(W;x_n)  e^{- y_n
		f(W;x_n)}.
	\end{equation}
	
	\noindent It follows that 
	\begin{equation}
	\dot{||W_k||^2} = 2 \sum_{n=1}^N y_n f(W;x_n)  e^{- y_n
		f(W;x_n)},
	\label{independent}
	\end{equation}
	\noindent which shows that the rate of growth of
	${||W_k||^2}$ is independent of $k$.  If we assume that
	$||W_1||=||W_2||=\cdots=||W_K||=\rho_1(t)$ initially, they will
	remain equal while growing throughout training. The norms of
	the layers are balanced, thus avoiding the situation in which
	one layer may contribute to decreasing loss by improving
	$\tilde{f}$ but another may achieve the same result by simply
	increasing its norm. 

	\subsection{Rate of growth of weights}\label{Weightgrowth}
	
	In linear 1-layer networks the dynamics of gradient descent yield
	$\rho \sim \log t$ asymptotically. For the validity of the results in
	the previous section, we need to show that the weights of a deep
	network also diverge at infinity. In general, the $K$ nonlinearly
	coupled equations  are not easily solved
	analytically. For simplicity of analysis, let us consider the case of
	a single training example $N=1$, as we expect the leading asymptotic
	behavior to be independent of $N$. In this regime we have
	\begin{equation}
	\rho_k\dot{\rho}_k = yf(V;x)\left(\prod_{i=1}^k\rho_i\right) e^{-\prod_{i=1}^K\rho_i y f(V;x)} 
	\end{equation} 
	Keeping all the layers independent makes it difficult to disentangle
	for example the behavior of the product of weights
	$\prod_{i=1}^K\rho_i$, as even in the 2-layer case the best we can do
	is to change variables to $r^2 = \rho^2_1 + \rho^2_2$ and
	$\gamma = e^{\rho_1 \rho^2 f_V(x)}$, for which we still get the
	coupled system
	\begin{equation}
	\dot{\gamma} = yf(V;x)^2 r^2, \qquad r\dot{r} = 2 \frac{\log\gamma}{\gamma},
	\end{equation}
	from which reading off the asymptotic behavior is nontrivial. 
	
\begin{figure*}[h!]\centering
		\includegraphics[width=1.0\textwidth]{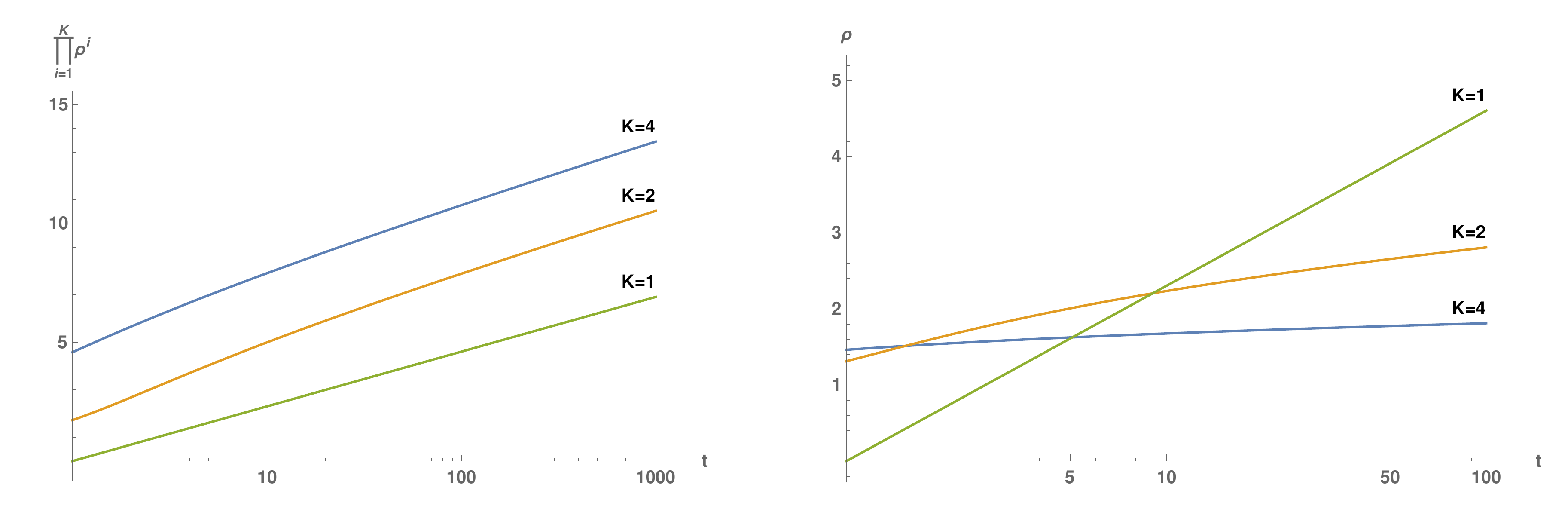}
		\caption{\it The left graph shows how the product of weights
			$\prod_{i=1}^K$ scales as the number of layers grows when
			running gradient descent with an exponential loss. In the
			1-layer case we have $\rho = \rho  \sim \log t$, whereas for
			deeper networks the product of norms grows faster than
			logarithmically. As we increase the number of
			layers, the individual weights at each layer diverge slower
			than in the 1-layer case, as seen on the right graph.}
		\label{rho_asymptotics}
	\end{figure*}

	We consider the case 
	$\rho := \rho_1 = \rho_2 = \ldots = \rho_k$. This turns out to be true
	in general (see Equation \ref{independent}). It gives us the single
	differential equation
	\begin{equation}
	\dot{\rho} = yf(V;x) K \rho^{K-1} e^{-\rho_k yf(V;x)}.
	\end{equation} 
	This implies that for the exponentiated product of weights we have
	\begin{equation}
	\left(e^{\rho_k yf(V;x)}\right)^{\dot{}} = f(V;x)^2 K^2 \rho^{2K-2}.
	\end{equation}
	Changing the variable to $R = e^{\rho_k yf_V(x)}$, we get finally
	\begin{equation}
	\dot{R} = f_V(x)^{\frac{2}{K}}K^2 \left(\log R\right)^{2 - \frac{2}{K}}.
	\end{equation}
	We can now readily check that for $K=1$ we get $R \sim t$, so
	$\rho \sim \log t$. It is also immediately clear that for $K>1$ the
	product of weights diverges faster than logarithmically. In the case
	of $K=2$ we get $R(t) = \textnormal{li}^{-1}(yf(V;x)K^2 t +C)$,
	where $\textnormal{li}(z) = \int_0^z dt / \log t$ is the logarithmic
	integral function. We show a comparison of the 1-layer and 2-layer
	behavior in the left graph in Figure \ref{rho_asymptotics}. For larger
	$K$ we get faster divergence, with the limit $K\to\infty$ given by
	$R(t) = \mathcal{L}^{-1}(\alpha_\infty t + C)$, where
	$\alpha_\infty = \lim_{K\to\infty} (yf(V;x))^{\frac{2}{K}}K^2$ and
	$\mathcal{L}(z) = \textnormal{li}(z) - \frac{z}{\log z}$.
	
	Interestingly, while the product of weights scales faster than
	logarithmically, the weights at each layer diverge slower than in the
	linear network case, as can be seen in the right graph in Figure
	\ref{rho_asymptotics}.
	
	\section{Critical points of the flow}
	\label{CriticalPoints}
	
	We discuss here in more details the stationary points of the gradient flow.
	
	The critical points of the flow induced by the Lagrange
	multiplier method described in section \ref{Lagrange}, $\dot{V}_k(t)=0$ are given by the
	following set of equations -- {\it as many as the number of weights}:

	\begin{equation}
	\dot{V_k}  =\rho \sum_n  e^{- \rho y_nf(V;x_n)} y_n\frac {\partial f(V;x_n)}
	{\partial V_k} - 2 \lambda_k(\rho)  V_k.
	\end{equation}
	
	The critical points of the unconstrained gradient system
	$\dot{V}_k(t)=0$ can be rewritten as (dropping here the subscript $k$ in
	$V_k$ for simplicity of notation)
	\begin{equation}
	0= \sum_n e^{-y_n f(x_n)}
	(I- V  V^T) \frac{\partial{f(V;x_n)}} {\partial
		V}.
	\end{equation}

        As it turns out, the evaluation of the Hessian of a linear
        un-normalized network is simple and shows that the global
        minima are hyperbolic and unique
        \cite{2017arXiv171010345S}\footnote{An easy calculation shows that
        Hessian wrt weights $w$ is
\begin{equation}
  H_{lin} = \sum_n^N x_n x^T_n e^{- x_n^T w},
\end{equation}
 which is a sum of outer products. If the data span the space, this leads to a non-degenerate Hessian.}

 For nonlinear, multilayer networks there are multiple minima for
 finite $\rho$ and it seems difficult to state that critical points of
 the flow -- after separability is reached -- are or are not hyperbolic minima.

 We argue however that some of them must be convex minima. The argument is
 as follows:

\begin{itemize}
\item After separability and after a single support vector (or group
  of SVs with same margin) dominates,  the margin increases until a
  critical point of the flow is reached (see lemma \ref{monotonic
    positive} below).
\item  If the rather weak conditions of
  \cite{pmlr-v49-lee16} are valid, then we can argue  that gradient descent
  (and especially  SGD) do not usually get stuck in saddle points. and
  that therefore an asymptotic critical point of the flow is likely to
  correspond to a convex minimum.

\end{itemize}

	\begin{lemma}
			If, after separation is reached, there is a
                        sufficiently large  $\rho$ for which the sum $\sum_{n=1}^N e^{-
				\rho y_nf(V;x_n)} \propto e^{-
				\rho y_*f(V;x_*)}$ then the  margin increases
			$y_*\frac{\partial f(V;x_*)} {\partial t}\ge 0$ (even if $\rho$ is
			kept fixed).
			\label{monotonic positive}
		\end{lemma}
		
		{\it Proof} $y_*f(V;x*)$ increases monotonically or is
		constant because
		\begin{equation}
		y_*\frac{\partial f(V;x*)}{\partial t}=\sum_k (\frac{\partial y_*f(V;x*)}{\partial
			V_k})^T \dot{V_k}=\sum_k \frac{\rho}{\rho_k^2}  e^{- \rho y_*f(V;x^*)} 
		\left(\left\lVert\frac{\partial{f(V;x^*)}} {\partial V_k}\right\rVert_F^2- {f(V;x^*)} ^2\right).
		\label{monotonic}
		\end{equation}
		
		\noindent Equation \ref{monotonic} implies
                $\frac{\partial y_* f(V,x_*)} {\partial t}\ge 0$ because
                $||f(V;x^*)||_F = ||V_k^T
                \frac{\partial{f(V;x^*)}} {\partial V_k}||_F \le
                ||\frac{\partial{f(V;x^*)}} {\partial V_k}||_F$,
                since the Frobenius norm is sub-multiplicative and
                $V_k$ has unit norm. We used Equation \ref{wdot2} that
                is
                $ \dot{V_k}=\frac{\rho}{\rho_k^2}\sum_{n=1}^N e^{-
                  \rho y_n f(V;x_n)} y_n
                (\frac{\partial{f(V;x_n)}} {\partial V_k}-V_k
                f(V;x_n))$.

                Notice that we do not have a proof that the stationary
                point can be realized -- in other words we do not know
                if $f$ can satisfy the equations
                $0= (I- V_k V_k^T) \frac{\partial{f(V;x^*)}}
                {\partial V_k}$ for all $k$. The equations are
                consistent however for ReLU networks as one can check
                by multiplying on the left by $V^T$.

%
%
%
%

%
%

\subsubsection{The Hessian has at least one positive eigenvalue}
	\label{CriticalPoints}
	
	We discuss here in more details the stationary points of the gradient flow.
	
	The critical points of the flow $\dot{V}_k(t)=0$ are given by the
	following set of equations -- {\it as many as the number of weights}:

	\begin{equation}
	\dot{V_k}  =\rho \sum_n  e^{- \rho y_nf(V;x_n)}y_n \frac {\partial f(V;x_n)}
	{\partial V_k} - 2 \lambda_k(\rho)  V_k.
	\end{equation}
	
	The critical points of the unconstrained gradient system
	$\dot{V}_k(t)=0$ can be rewritten as (dropping here the subscript $k$ in
	$V_k$ for simplicity of notation)
	\begin{equation}
\sum_n e^{-y_n f(x_n)}y_n
	(I- V  V^T) \frac{\partial f(V;x_n)} {\partial
		V}.
	\end{equation}

The equations give the same zeros since they only differ by a positive
factor $\rho$ -- that we assume fixed here.

We consider the Hessian of $L$ associated with the dynamical system
$\dot{V_k} = - F(V_k,\rho_k) = - \nabla_{V_k} L$ w.r.t. $V_k$. The Hessian tells us about the linearized dynamics
around the asymptotic critical point of the gradient and thus about
the sufficient conditions for local minima under the norm
constraint. The question is whether $-H$ is negative semidefinite or
negative definite. The second case is needed to have {\it sufficient
  conditions} for local minima. We consider the quadratic form
$\sum^{K,K}_{k,{k'}}V_k^T H V_{k'}$, where $V_k$ correspond to
critical points of the gradient (the Hessian for the unconstrained and
the constrained system differ by a factor $\rho$). This represent one
specific direction only. We will show that the Heassian has positive
eigenvalues for $v$

	The negative Hessian for the standard gradient descent case is 
	\begin{equation}
	-H^{i,j}_{k, {k'}}= \sum_n e^{-y_n \rho f(V;x_n)}  (A_n + B_n + C_n)
	\end{equation}
	
	\noindent with 
	
	$A_n= (-\rho  \frac {\partial f (V;x_n)}
	{\partial V_k^i} (I-V_{k'}^j(V_{k'}^T)^\ell) \frac{\partial{f(V;x_n)}} {\partial
		V_{k'}^\ell})$
	
	\noindent  and
	
	$B_n= y_n\delta_{k k'}
	( -(\delta_{i,j} V_{k'}^\ell + \delta_{\ell, i} V_{k'}^i) {\frac {\partial f(V;x_n)}
		{\partial V_{k'}^\ell}})$
	
	\noindent  and 
	
	$C_n= (I-V_{k'}^j(V_{k'}^T)^\ell)  \frac {\partial^2 f (V;x_n)}
	{\partial V_k^i \partial V_{k'}^\ell}$.
	
For the form  $\sum^{K,K}_{k,k'}V^i_k (-H^{i,j}_{k, {k'}} ) V^j_{k'}$ we find
	
	\begin{itemize}
		
		\item the first term disappears, because $(I-V_{k'}^j(V_{k'}^T)^\ell) V_{k'}^j = 0$ (because $ (I-V_{k'}^j(V_{k'}^T)^\ell)  = S$ and $S V = 0$).
		\item the second term gives $ - 2 K \rho \sum_n e^{- \rho
			y_nf(V;x_n)}  y_nf(V;x_n)$
		\item the third term disappears for the same reason as the first one.
		
	\end{itemize}
	
	{\it Thus if gradient descent separates the data at $T_0$, for $t>T_0$
		$-H$ is negative definite for any finite $\rho$ in the
                $v$ direction. It
		becomes negative semi-definite in the limit $\rho = \infty$. }.

	{\bf notice that convergence at finite time probably means a
          small number of support vectors with same value and so good
          generalization...add theorem!}

\subsection{New algorithms?}
An examination of the gradient descent dynamics on exponential losses,
discussed above, suggests experimenting with {\it a new family of
  algorithms} in which $\rho(t)$ is designed and controlled
independently of the gradient descent on the $V_k$: we need $\rho$ to
grow in order to drive the exponential loss towards the ideal
classification loss but our real interest is in maximizing the margin
of $f(V;x)$ . The ideal time course of $\rho(t)$ could be made
adaptive to the data. It seems possible to test dynamics that may be
quite different from $\rho \propto \log t$. Of course the role of
$\rho$ is complex: in the initial stages of GD it makes sense that the
best solutions are solutions with $||V_k||=1$ and lowest exponential
loss.  Intuitively they correspond to solutions that separate and have
minimum $||W_k||$.

	\section{Remark}
	\label{Connection}

		 In the linear one-layer case
		$f(V;x)=v^T x$, the stationary points of the gradient are given
		by $\sum \alpha_n\rho(t) (x_n - v v^T x_n).$ The Lagrange multiplier
		case is similar, giving $\sum \alpha_n\rho(t) x_n - \lambda v=0$, with
		$\lambda$ satisfying $\lambda^2= \sum_n e^{2\rho y_nv^T x_n}\rho^2 y_nx_n
		$. Thus $\lambda \to 0$ for $\rho \to \infty$.

	\section{$L_p$ norm constraint}
	
	To understand whether there exists an implicit complexity control in
	standard gradient descent of the weight directions, we  check whether there exists an
	$L_p$ norm for which unconstrained normalization is equivalent to
	constrained normalization.
	
	From Theorem \ref{Theorem1} we expect
	the constrained case to be given by the action of the following
	projector onto the tangent space:
	\begin{equation}
	S_{p} = I-\frac{\nu \nu^T}{||\nu||_2^2} \quad\textnormal{with}\quad \nu_i=\frac{\partial ||w||_p}{\partial w_i} = \textnormal{sign}(w_i)\circ\left(\frac{|w_i|}{||w||_p}\right)^{p-1}.   
	\end{equation}
	The constrained Gradient Descent is then
	\begin{equation}
	\dot{\rho_k}= V^T_k \dot{W_k}  \quad
	\dot{V_k} = \rho_k S_p \dot{W_k}.
	\label{ConstrainedGradP}
	\end{equation}
	
	On the other hand,  reparameterization of
	the unconstrained dynamics in the $p$-norm gives (following Equations \ref{rhodot} and \ref{vdot})
	\begin{equation}
	\begin{split}
	\dot{\rho_k}&= \frac{\partial ||W_k||_p}{\partial W_k} \frac{\partial
		W_k}{\partial t}= \textnormal{sign}(W_k)\circ\left(\frac{|W_k|}{||W_k||_p}\right)^{p-1} \cdot \dot{W_k} \\
	\dot{V_k}&= \frac{\partial V_k}{\partial W_k} \frac{\partial
		W_k}{\partial t}= \frac {I - \textnormal{sign}(W_k) \circ\left(\frac{|W_k|}{||W_k||_p}\right)^{p-1}W_k^T}{||W_k||_p^{p-1}} \dot{W_k}.
	\end{split}
	\end{equation}
	These two dynamical systems are clearly different for generic
	$p$ reflecting the presence or absence of a regularization-like
	constraint on the dynamics of $V_k$.
	
	As we have seen however, for $p=2$ the 1-layer dynamical system obtained by
	minimizing $L$ in $\rho_k$ and $V_k$ with $W_k=\rho_k V_k$ under the constraint
	$||V_k||_2=1$, is the weight normalization dynamics

	\begin{equation}
	\dot{\rho_k}=V_k^T \dot{W_k} \quad
	\dot{V_k}= S \rho_k \dot{W_k} ,
	\label{WN}
	\end{equation}
	
	\noindent which is quite similar to the standard gradient
	equations 
	\begin{equation}
	\dot{\rho_k}= V_k^T \dot{W_k}  \quad
	\dot{v} =\frac{S}{\rho_k} \dot{W_k}.
	\label{StandardGrad}
	\end{equation}

	\section{Linear networks: rates of convergence}
	\label{LinearNets} 
	We consider here the linear case. We rederive some of the results by
	\cite{2017arXiv171010345S} in our gradient flow framework.
	In the separable case of a linear network ($f(x)=\rho v^T x$) the dynamics is
	\begin{equation}
	\dot{\rho}=  \frac{1}{\rho} \sum_{n=1}^N  e^{-
		\rho y_nv^Tx_n} y_nv^Tx_n
	\end{equation}
	
	\noindent and 
	
	\begin{equation}
	\dot{v}=\frac{1}{\rho}\sum_{n=1}^N  e^{- \rho y_nv^T x_n} y_n
	(x_n- v  v^T x_n).
	\end{equation}
	
	As discussed earlier there are $K$ support vectors with the same
	smallest margin. For $t$ increasing, since $\rho \approx \log t$,
	
	\begin{equation}
	\dot{v} \propto \frac{1}{\rho} \sum_{j=1}^K \alpha_j
	(I- vv^T) x_j.
	\label{T}
	\end{equation}
	\noindent with $\alpha_j=y_je^{-\rho y_jv^T x_j}$. If gradient descent
	converges, the solution $v$ satisfies $v v^T x = x$, where
	$x= \sum_{j=1}^K \alpha^*_j x_j$. It is easy to see that
	$v= ||x||^2 x^\dagger$ -- where $x^\dagger$ is the pseudoinverse of
	$x$ -- is a solution since the pseudoinverse of a non-zero vector $z$
	is $z^\dagger= \frac{z^T}{||z||^2}$. On the other hand, in the linear
	case the operator $T$ in $v(t+1)=T v(t)$ associated with equation
	\ref{T} is not expanding because $v$ has unit norm. Thus
	\cite{Ferreira1996} there is a fixed point $v=x$ which is {\it
		independent of initial conditions}. 

	To simplify the analysis we assume in the following that there is a
	single effective support vector (that is a set of points with the same
	margin). As one of the figures show, this is often not correct. The
	rates of convergence of the solutions $\rho(t)$ and $v(t)$, derived in
	different way in \cite{2017arXiv171010345S}, may be read out from the
	equations for $\rho$ and $v$. It is easy to check that a general
	solution for $\rho$ is of the form $\rho \propto C \log t$. A similar
	estimate for the exponential term, gives
	$e^{- \rho y_nv^T x_n} \propto \frac{1}{t}$. We claim that a solution for
	the error $\epsilon= v-x$, since $v$ converges to $x$, behaves as
	$\frac{1}{\log t}$.  In fact we write $v =x+\epsilon$ and plug it in
	the equation for $v$ in \ref{T} (we also assume for notation
	convenience a single data point $x$). We obtain
	\begin{equation}
	\dot{\epsilon}=\frac{1}{\rho}  e^{- \rho v^T x} (x- (x+\epsilon)
	(x+\epsilon)^T x) = \frac{1}{\rho}  e^{- \rho v^T x} ( x- x - x
	\epsilon^T - \epsilon x^T)
	\label{TT}
	\end{equation}
	\noindent which has the form
	$\dot{\epsilon}=-\frac{1}{t \log t} (2 x \epsilon^T)$. A solution of
	the form $\epsilon \propto \frac{1}{\log t}$ satisfies the equation:
	$-\frac{1}{t \log^2 t }= -B \frac{1}{t \log^2 t}$. Thus the error
	indeed converges as $\epsilon \propto \frac{1}{\log t}$.
	
\begin{figure*}[h!]\centering
		\includegraphics[trim = 80 450 80 80,width=1.0\textwidth, clip]{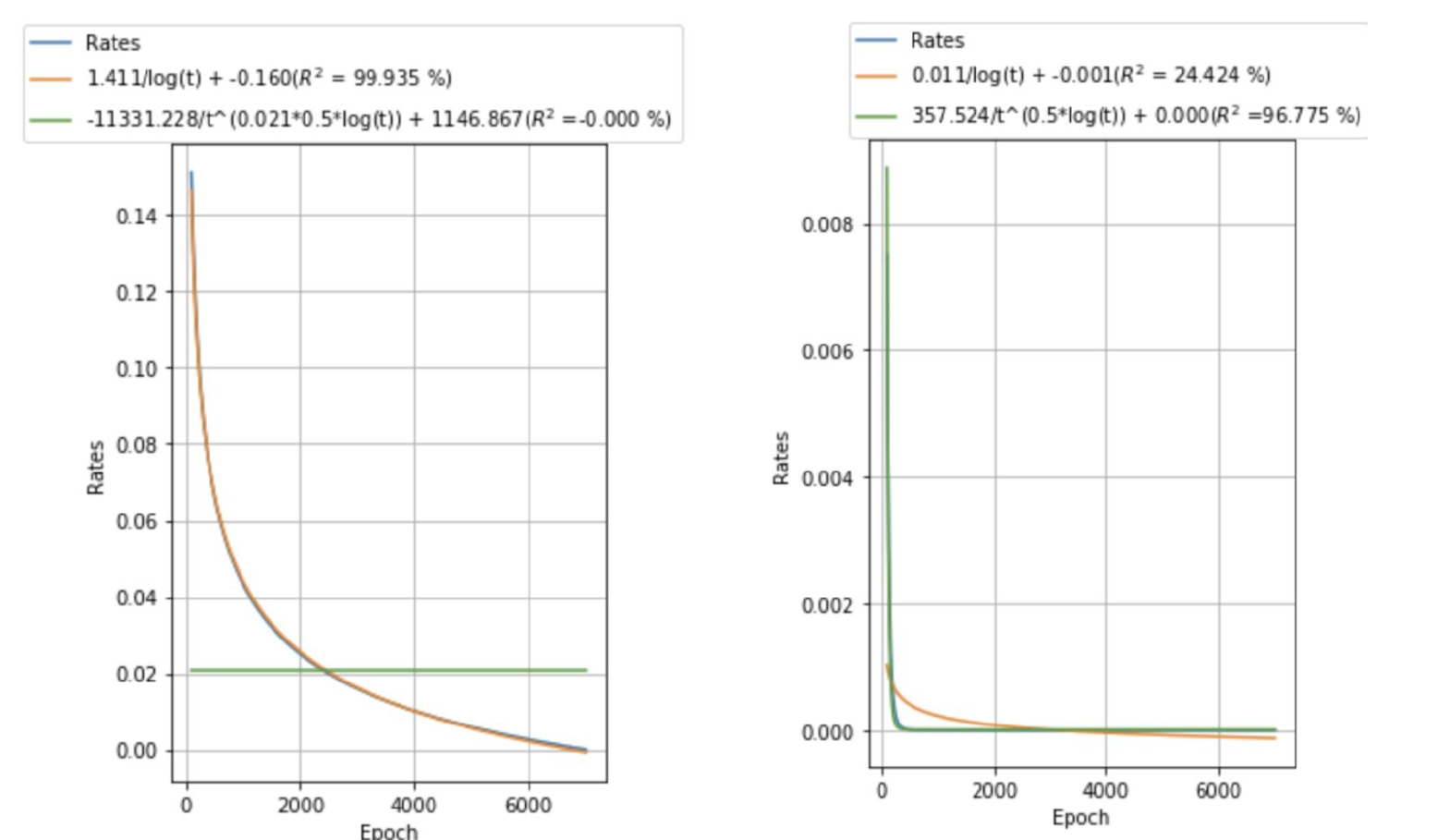}
		\caption{\it Rates of convergence of $v(t)$ for a linear
			network trained on 1 training example on the IRIS
			dataset. Left: Standard gradient descent converges in
			direction at the rate $\mathcal{O}(\frac{1}{\log
				t})$. Right: Weight Normalization changes the convergence
			rate to that of $\mathcal{O}(t^{-1/2 \log t})$. We obtain
			more complex behavior in the presence of multiple support
			vectors with different margins.}
		\label{convergencerates}
	\end{figure*}

\begin{figure}
		\centering
		\begin{minipage}{.5\textwidth}
			\centering
			\includegraphics[width=.99\linewidth]{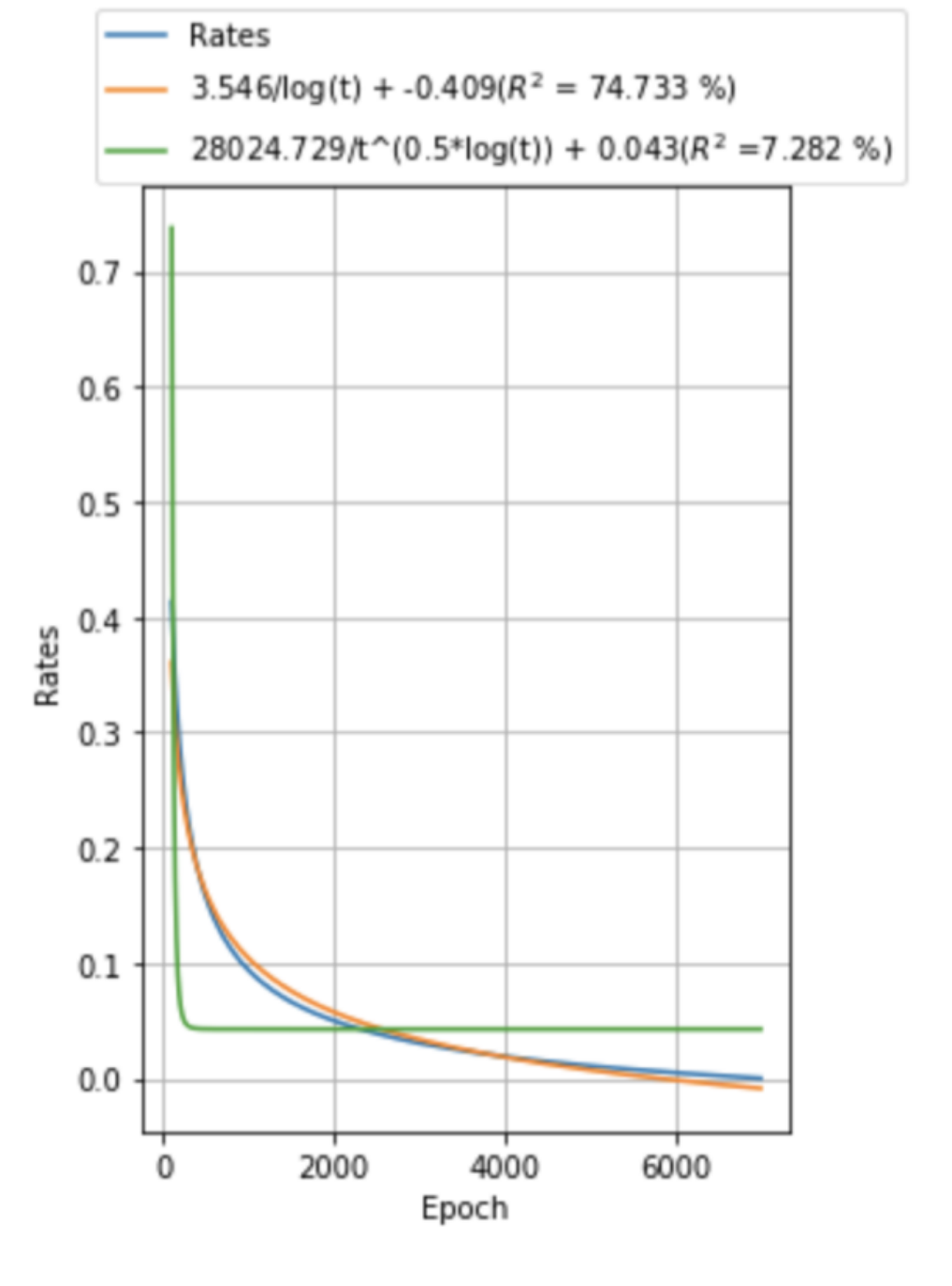}
		\end{minipage}%
		\begin{minipage}{.5\textwidth}
			\centering
			\includegraphics[width=.99\linewidth]{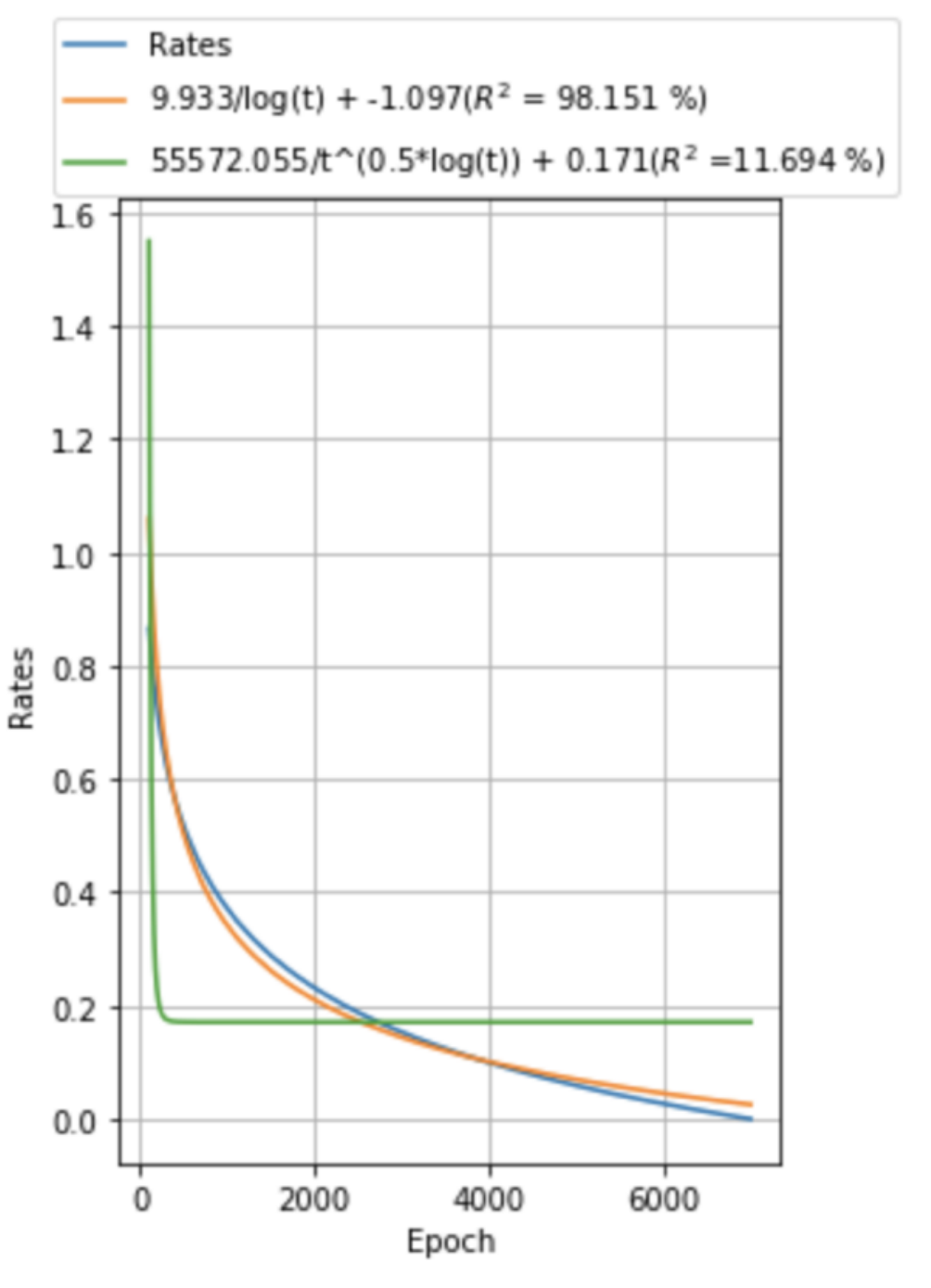}
		\end{minipage}
		\caption{\it Rates of convergence of $v(t)$ for a linear
			network trained on all the training examples on the IRIS
			dataset. Left: Standard gradient descent. Right: Weight Normalization. The goodness of fit in the both cases drops compared to the 1 example case. In the case of WN, $\mathcal{O}(\frac{1}{\log
				t})$ seems like a much better fit. Multiple data points imply more complicated dynamics and a slower convergence to the support on a small subset of inputs.}
		\label{convergencerates2}
	\end{figure}

	A similar analysis for the weight normalization equations 
	considers the same dynamical system with a change in the equation for
	$v$ which becomes
	
	\begin{equation}
	\dot{v} \propto e^{-\rho}  \rho  (I- v v^T) x.
	\label{T-WN}
	\end{equation}
	This equation differs by a factor $\rho^2$ from equation \ref{TT}. As
	a consequence equation \ref{T-WN} is of the form
	$\dot{\epsilon}=-\frac{\log t}{t} \epsilon$, with a general solution
	of the form $\epsilon \propto t^{-1/2\log t}$. Numerical experiments
	confirm these rates for linear networks with one data point, see
	Figure \ref{convergencerates}, but suggest modifications in the case
	with $N\neq 1$, see Figure \ref{convergencerates2}. In summary, {\it GD with weight normalization
		converges faster to the same equilibrium than standard gradient
		descent: the rate for $\epsilon= v- x$ is $\frac{1} {t^{1/2\log t}}$
		vs $\frac{1}{\log t}$.}
	
	\begin{figure}
		\centering
		\begin{minipage}{.5\textwidth}
			\centering
			\includegraphics[trim = 120 100 120 80, width=.99\linewidth ,clip ]{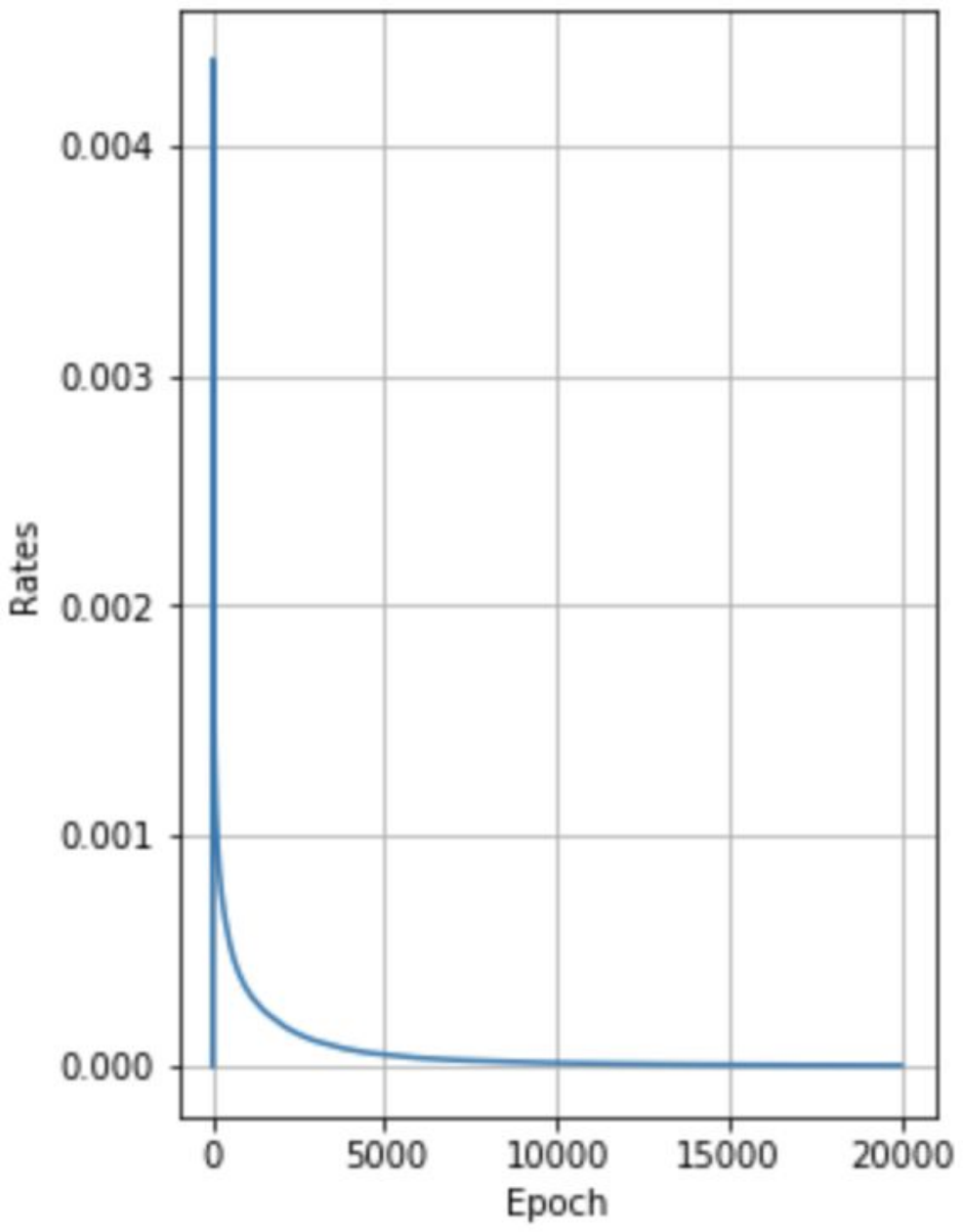}
		\end{minipage}%
		\begin{minipage}{.5\textwidth}
			\centering
			\includegraphics[trim = 120 100 120 80, width=.99\linewidth, clip]{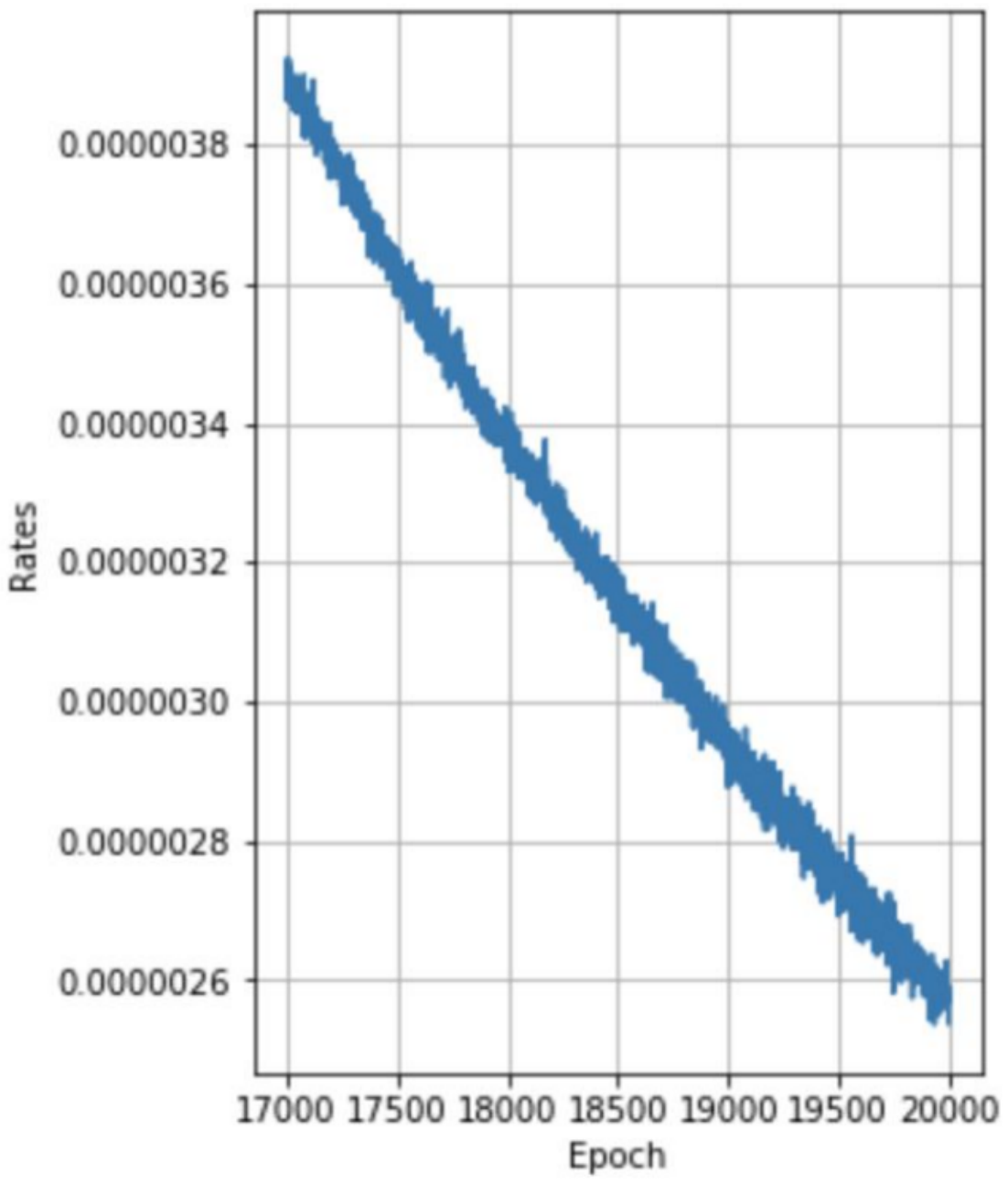}
		\end{minipage}
		\caption{\it Rates of convergence of $v(t)$ compared at every epoch for a linear
			network trained on all the training examples on the IRIS
			dataset. Left: All Epochs. Right: Last
			3000 epochs. Towards convergence, there are no oscillations in the rates. The oscillating behavior observed in the last epochs are a result of numerical error. }
		\label{convergenceratesoscillations}
	\end{figure}

	Our simplified analysis implies that batch normalization has the same convergence
	rate as weight normalization (in the linear one layer case BN is
	identical to WN). As we discussed, it is however different wrt WN in
	the multilayer case: it has for instance separate normalization for
	each unit, that is for each row of the weight matrix at each layer.

\section{Scaling $\rho$}
	
\begin{figure}
		\centering
		\begin{minipage}{.5\textwidth}
			\centering
			\includegraphics[trim = 80 140 80 80, width=.99\linewidth, clip]{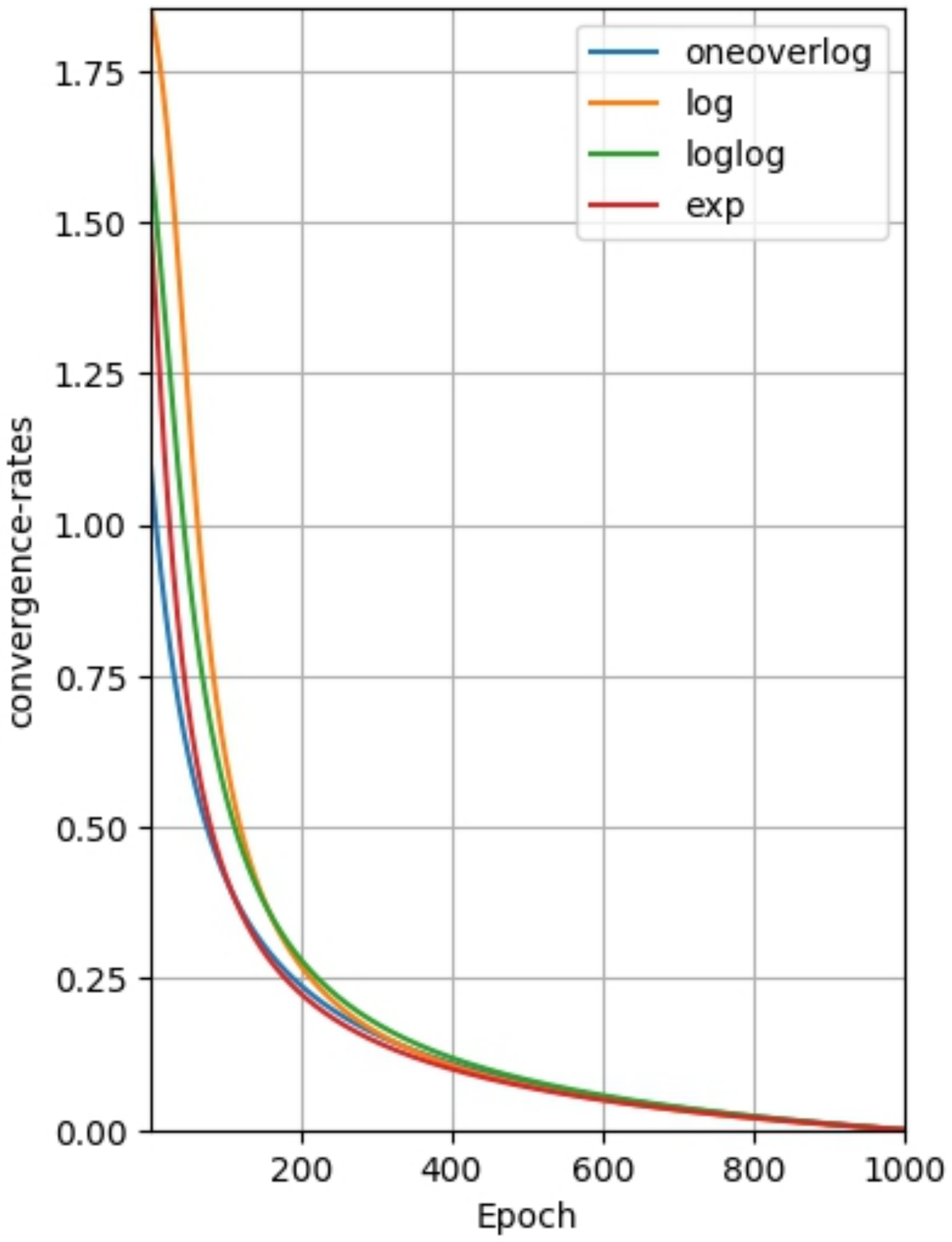}
		\end{minipage}%
		\begin{minipage}{.70\textwidth}
			\centering
			\includegraphics[trim = 80 150 80 130, width=.99\linewidth, clip]{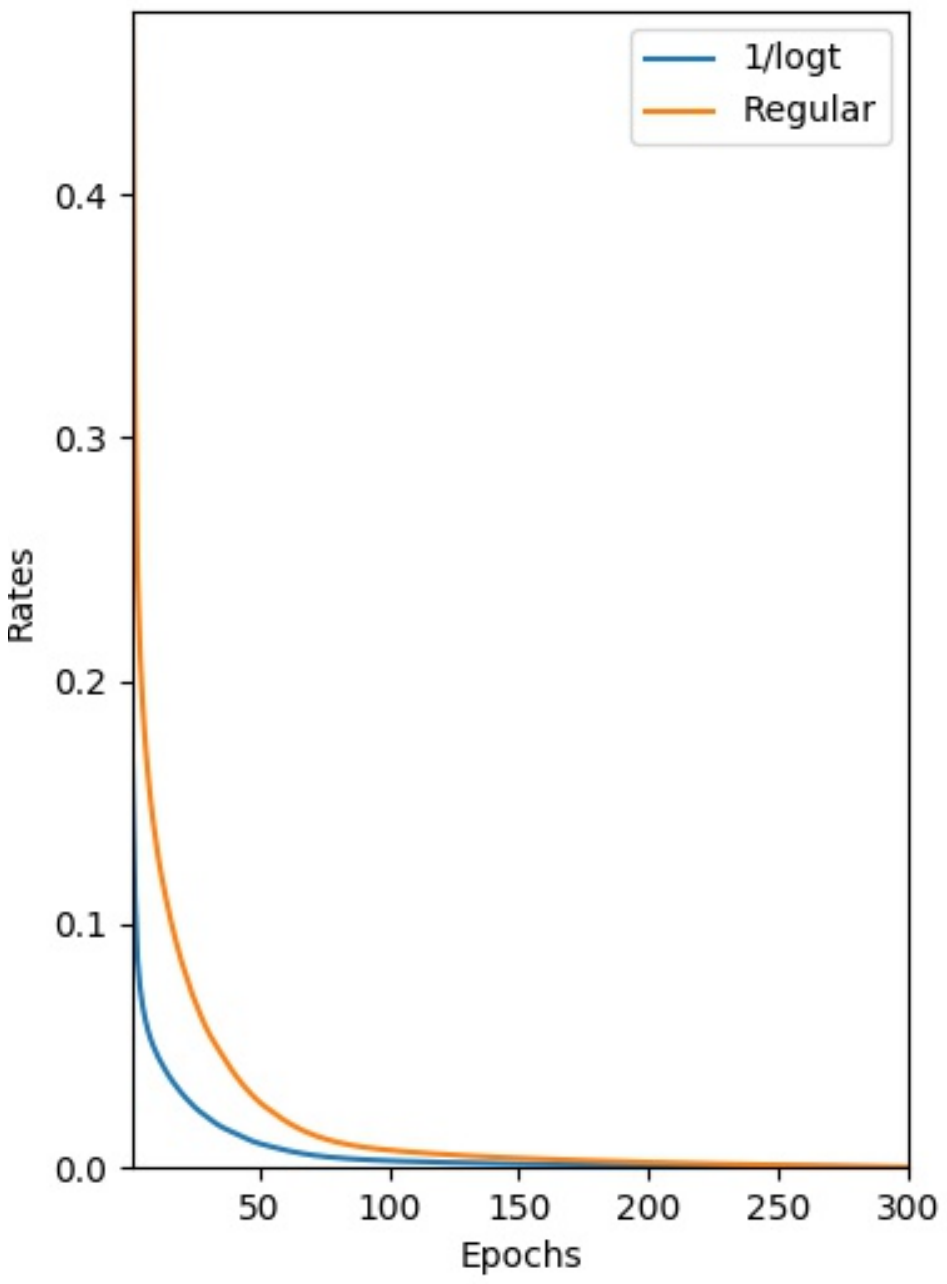}
		\end{minipage}
		\caption{\it Rates of convergence of $v(t)$ with $\rho$ set as several functions of $t$ for a linear
			network trained on all the training examples on the IRIS
			dataset.
			Left: Convergence rates when $\rho$ was set to different functions of $t$. Right: Rates of convergence with $rho$ set to $\frac{1}{\log{t}}$ compared to the original rates of convergence. }
					\label{convergenceratesrhoasfunctions}.
	\end{figure}

	With the observation that weight normalization leads to faster rates of convergence than unconstrained dynamics, it is natural to consider alternative scalings of  $\rho$. In the dynamics of $\dot{V}$, We experiment with different rescalings $\alpha(t)$. The dynamics of normalized weights then become
	
	\begin{equation}
	\dot{V}=\alpha\frac{1}{\rho}\sum_{n=1}^N  e^{- \alpha\rho y_nV^T x_n} y_n
	(x_n- V  V^T x_n).
	\end{equation}
	Specificially, we will consider $\alpha(t) = \frac{1}{\log{(t)}}$, $\log{(t)}$, $\log{\log{(t)}}$, $\exp{(t)}$. See Figure \ref{convergenceratesrhoasfunctions}
	
	Additionally, to better understand the effects of $\rho$ on the test loss and error, we run experiments similar to \cite{DBLP:journals/corr/abs-1807-09659} where networks were initialized with weights of different Gaussian standard deviations and a linear relationship between test loss and error emerges. Here, we scale $\rho$ linearly to observe when this linear relationship disappears as $\rho$ is increased.

	\begin{figure}
		\centering
		\begin{minipage}{1\textwidth}
			\centering
			\includegraphics[trim = 0 40 0 40, width=1.1\linewidth, clip]{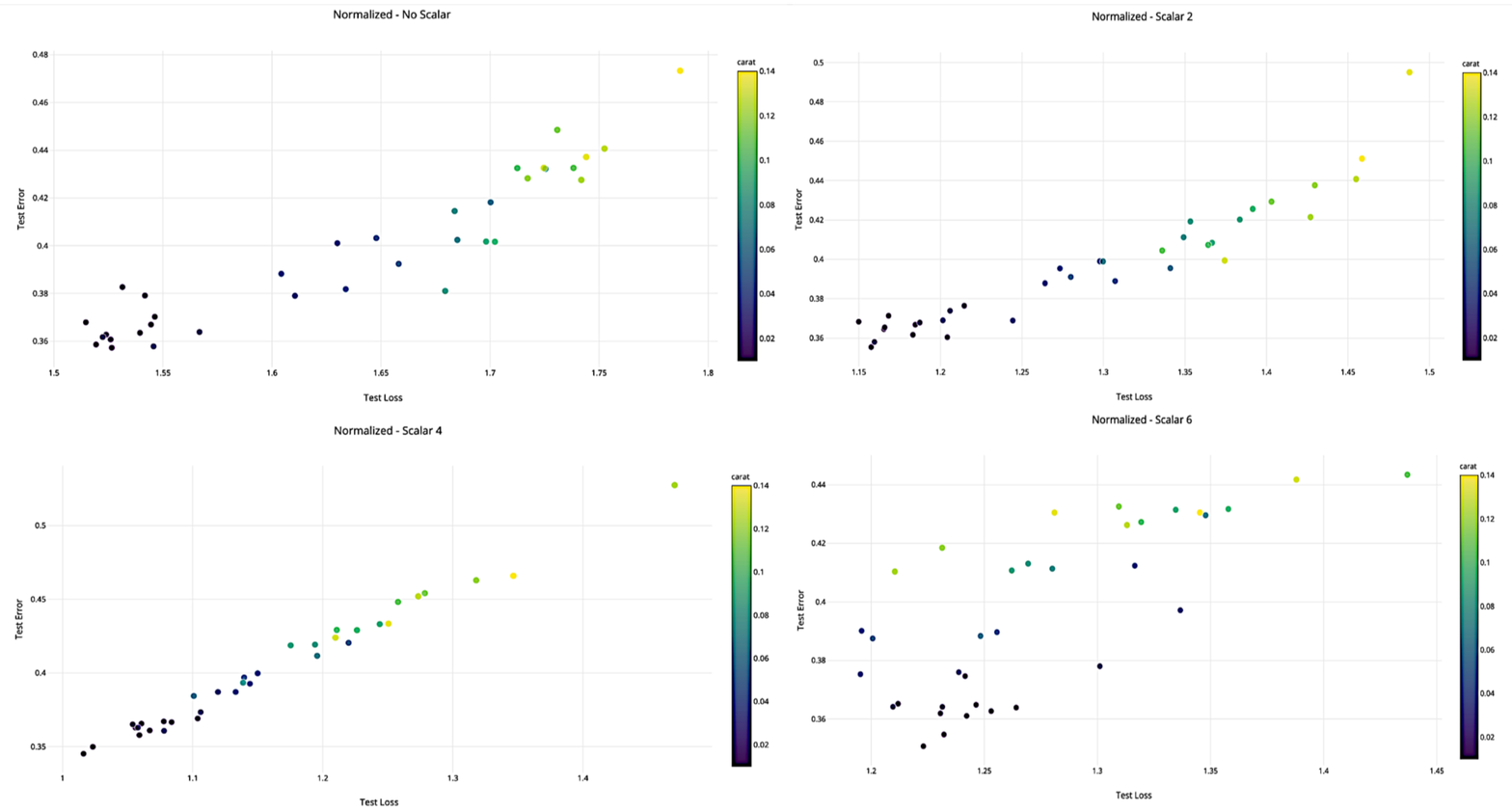}
		\end{minipage}%
		\caption{\it The experiments are measured as in Figure \ref{main}, here we plot the normalized test loss vs. test error. As we scale the normalized network, the linear relationship disappears around $\rho \approx 4$. }
		\label{weight1}
	\end{figure}

\begin{figure}
	\centering
	\begin{minipage}{1\textwidth}
		\centering
		\includegraphics[trim = 0 0 0 0, width=1\linewidth, clip]{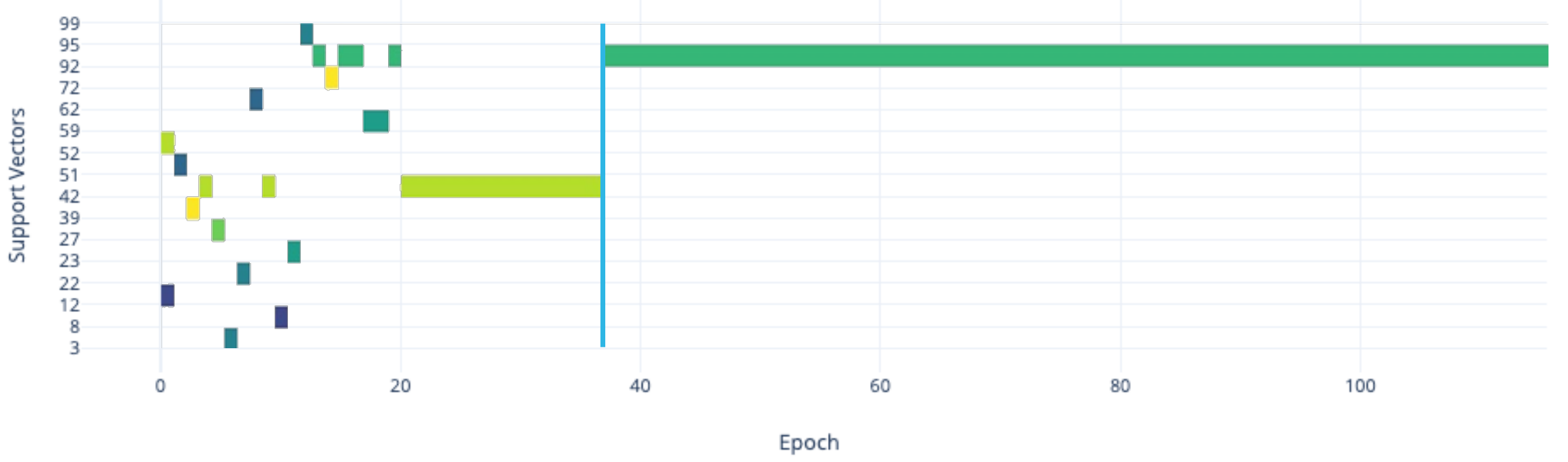}
	\end{minipage}%
	\caption{\it A 6-layer neural network implemented in PyTorch was trained  on a subset of 100 datapoints from CIFAR-10 with Gradient Descent
		(GD) on cross-entropy loss. Support vectors were obtained by taking the smallest of the distances between the 
		class output and the second highest class output for each datapoint, $\operatorname*{argmin}( f_{y_{i}} - \max\limits_{j \neq i}f_{y_{j}})$.
		The blue line represents data separation (100 $\%$ accuracy training). The network has 4 convolutional layers (filter size 3 $\times$ 3, stride 2) 
		and two fully-connected layers. Each convolutional layer is followed by batch-normalization. The number of feature maps (i.e., channels) in hidden layers are 16, 32, 64, 128 respectively. In total, the network has 6,400 parameters. The learning rate
		was constant throughout and set to 0.1 and momentum to 0.9. Neither data augmentation nor regularization is performed.  }
	\label{weight2}
\end{figure}

\begin{figure}
	\centering
	\begin{minipage}{1\textwidth}
		\centering
		\includegraphics[trim = 0 0 0 0, width=1\linewidth, clip]{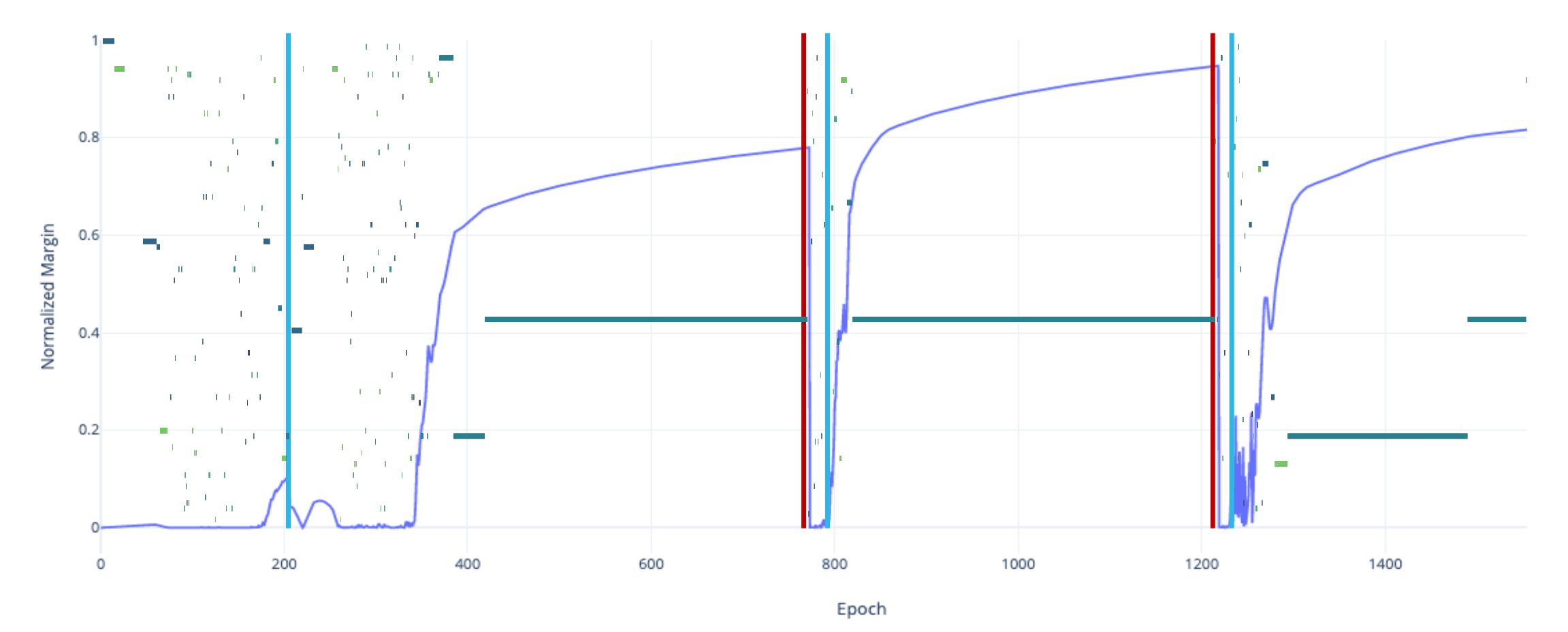}
	\end{minipage}%
	\caption{\it A convolutional neural network with same parameters and similar architecture as the one in Figure 9, with the distinction of no batch-normalization, was trained on 100 data points from the CIFAR10 dataset. 
		The margin $\operatorname*{argmin}( f_{y_{i}} - \max\limits_{j \neq i}f_{y_{j}})$ and the data points closest to the margin solution (support vectors) at each epoch are shown.			
		The network separates the data (reaches $100\%$ training accuracy) at epoch 200 represented by the first blue line. 
		After the rate of change of the margin is $\leq 0.00002$ (margin convergence) perturbations to the network were performed with respect to the gradient of each individual weight, $0.5\left \| \nabla_{w}  L \right \|$.
		Each perturbation was performed after the network separates the data again (blue lines) and reaches margin convergence as indicated by the red lines. 
		The network converges to one support vector and then jumps between minima after each perturbation but eventually comes back to the same support vector. }
	\label{weight3}
\end{figure}
	
\section{Epilogue}

The first observation is that $y=Wx$ can be made better conditioned by
doing $y=ABx$ Suppose $W$ is $n,n$; then $A$ could be $n,d$
and $B$ $d,n$ with $AB=W$ which means $A=WB^\dagger$ or $B=A^\dagger
W$. Can I have better condition numbers with $AB$?

				
\end{document}